\RequirePackage{fix-cm} 
\documentclass[10pt,twocolumn,letterpaper]{article}
\usepackage[T1]{fontenc}
\usepackage{cvpr}
\usepackage{graphicx}
\usepackage{pifont}
\usepackage{tikz}
\usepackage{listings}
\usetikzlibrary{arrows.meta,calc}

\definecolor{cvprblue}{rgb}{0.21,0.49,0.74}
\definecolor{projectlink}{RGB}{212,31,134}
\usepackage[pagebackref,breaklinks,colorlinks,allcolors=cvprblue]{hyperref}

\def\confName{CVPR}
\def\confYear{2027}
\title{Support2Sim: Visually Faithful Real2Sim with Stable Object Supports}
\author{%
  Samarth Chopra\textsuperscript{%
    \href{mailto:csamarth@engineering.upenn.edu}{\textcolor{black}{\ding{41}}}},
  Ankit Prabhu, Akshayan Nagarajah\\
  Sumer Murthy, Pratik Chaudhari, Vijay Kumar\\
  University of Pennsylvania\\
  {\small\texttt{\{csamarth, praankit, aknag, sumerm, pratikac, kumar\}@engineering.upenn.edu}}%
}
\date{November 2026}

\makeatletter
\apptocmd{\@maketitle}{\par\noindent
\begin{minipage}{\textwidth}
  \centering
  \captionsetup{type=figure}
  \includegraphics[width=.95\linewidth]{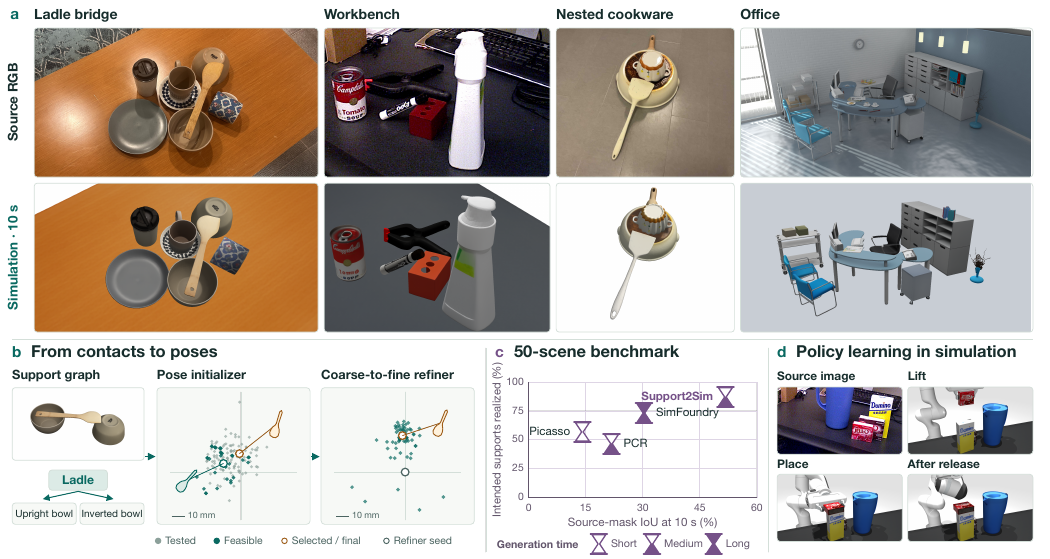}
  \caption{\textbf{Overview of Support2Sim.}
  (a) Support2Sim produces visually faithful scene reconstructions in simulation,
  shown beside their source images as matched-camera renders after 10 seconds.
  (b) Contact-guided pose initialization and coarse-to-fine refinement enable
  fine-grained pose optimization, illustrated by the ladle's support graph
  and pose proposals.
  (c) Mean source-mask intersection-over-union (IoU) at 10 seconds and
   the number of supports realized over 50 scenes. Hourglass correspond to the median generation times, including shared preprocessing: short $<10$ min,
  medium $10$ to $<20$ min, and long $\ge20$ min.
  (d) Learned box-stacking policy in a reconstructed scene.}
  \label{fig:overview}
\end{minipage}
\par
}{}{%
  \PackageError{sceneanything}{Could not place the overview under the title}{}}
\makeatother

\begin{document}
\maketitle
\begingroup
\renewcommand{\thefootnote}{}
\footnotetext{This work was supported by funding from the IoT4Ag
Engineering Research Center under NSF Cooperative Agreement EEC-1941529
and from the ARL DCIST CRA under Agreement W911NF-17-2-0181.}
\endgroup
\begin{abstract}
Reconstructing a scene for physics simulation requires recovering its spatial
layout and the support relationships that govern its behavior. We present
Support2Sim, a Real2Sim system that reconstructs simulation-ready scenes
from a single RGB image. Our design objective is to maximize source fidelity,
support realization, and stability within a practical time budget for scene
generation. Support2Sim infers a support graph, which a contact-guided pose
initializer uses to propose poses for objects with multiple supports. The
initializer generates and ranks candidates by geometric fit and
source-image agreement. A coarse-to-fine refiner allocates a fixed search
budget across supported objects and evaluates pose updates with physics rollouts.
After an accepted update, it searches from the new pose at a finer translation
and rotation scale. We benchmark our method on 50 scenes spanning four
datasets. Compared with the strongest baseline,
Support2Sim improves source-mask IoU after 10 seconds by $21.4$\,pp,
realizes $87.1\%$ versus $73.2\%$ of intended supports, and reduces mean
object drift from $79.6$ to $45.3$\,mm.
See our project page at
{\hypersetup{urlcolor=projectlink}\url{https://support2sim.github.io/}}
for more details.
\end{abstract}

\section{Introduction}
\label{sec:introduction}

Training general-purpose robot policies at scale requires interaction
data across diverse objects, tasks, and
environments~\cite{openx2024embodiment,khazatsky2024droid,walke2023bridgedata,black2025pi0}. Real-to-sim
reconstruction creates reusable simulated scenes for generating demonstrations
to supplement real-world robot data or training policies directly in
simulation~\cite{tornevillasevil2024rialto,nasiriany2026robocasa365,ranawaka2026simfoundry,liu2026worldgrounding,deshpande2026molmobot,escontrela2025gaussgym}.
These scenes also enable repeatable policy testing before deployment,
reducing reliance on costly hardware setup and resets. Evaluations in simulation
and on real robots show that policy performance in reconstructed scenes
can track real-world performance~\cite{li2025simpler,jain2025polaris,zhang2025softbodyeval,genesis2026genesisworld1}.

For simulated rollouts to reproduce a real task, the scene must preserve
the visual cues and physical interactions on which the policy
relies~\cite{wang2025embodiedreamer,liu2026worldgrounding,li2025simpler}.
\par
\smallskip
\noindent\phantomsection\label{obj:system-design}%
\textbf{System Design Objective}\par
\nopagebreak[4]
\smallskip
\begingroup
\definecolor{objectiveFill}{HTML}{F2F7F5}
\definecolor{objectiveBorder}{HTML}{9BBEB4}
\noindent\begin{tikzpicture}
  \node[draw=objectiveBorder, fill=objectiveFill, rounded corners=4pt,
    line width=0.5pt, inner sep=5pt, outer sep=0pt,
    text width=\dimexpr\linewidth-10.5pt\relax, font=\small] {%
    {\centering\color{black}\boldmath
    \(\displaystyle
      \begin{aligned}
        \operatorname*{maximize}_{S}\,&
        \bigl(\text{\textbf{source fidelity}},\,
              \text{\textbf{support realization}},\,
              \text{\textbf{stability}}\bigr)\\
        \text{\textbf{subject to}}\,& T_{\mathrm{gen}}(S) \le B .
      \end{aligned}
    \)\par}
    \vspace{3pt}
    $T_{\mathrm{gen}}(S)$ is the wall-clock time to generate scene $S$,
    and $B$ is the available per-scene time budget.
  };
\end{tikzpicture}
\par
\endgroup
\smallskip
\noindent
\emph{Source fidelity} preserves the visible object arrangement used by a
vision-based policy. \emph{Support realization} measures sustained
force-bearing contact between object pairs linked by the inferred support
graph, without requiring the original contact locations or force
distribution. \emph{Stability} keeps objects near their reconstructed poses
after release under gravity. For the ladle spanning two bowls in
Fig.~\ref{fig:overview}, a gap at one end can cause rotation during
settling, shifting the handle before a grasp. Such shifts alter starting
states and make evaluation failures harder to attribute to the policy.

These requirements are difficult to satisfy together from one RGB image.
Image-to-3D methods can recover object meshes and estimate their
layout~\cite{chen2026sam3d,meng2026scenegen}, yet different 3D poses can
project to similar silhouettes. Small errors in depth or pose can
therefore create gaps or overlaps between surfaces that should
touch~\cite{yu2026picasso,li2026phiscene,xiang2026physicsreal2sim}.
Object poses are coupled: moving a support to close one gap can disturb
another contact, and settling can shift the scene away from the observed
layout. Optimizing object poses, physical properties, or shape can be time
intensive~\cite{huang2026sparcs,xiang2026physicsreal2sim}, motivating bounded
search budgets for scalable Real2Sim.
Agentic harnesses show promise for automated scene construction, but
their inference costs can become prohibitive at
scale~\cite{tsui2026frontieragents,pfaff2026scenesmith}

We present \textbf{Support2Sim}, a real-to-sim system that reconstructs
rigid-body scenes from a single RGB image. An inferred support graph
specifies the intended resting relationships, providing geometric targets
for constructing the layout and support relationships to assess in simulation.

\begin{enumerate}[nosep]
  \item \textbf{Support-guided real-to-sim.} We combine a support graph
  with sampling-based pose initialization. The graph determines seating
  order and guides the search for poses compatible with multiple inferred
  supports. We maximize source-image agreement among the
  feasible rendered samples to preserve source fidelity.
  \item \textbf{Coarse-to-fine pose refinement.} Within a fixed search
  budget, we test pose updates in short physics rollouts and prioritize
  support groups using support deficits, image fit, and motion. After
  accepting an update, we search from the new pose at a finer scale.
  \item \textbf{Evaluation across indoor datasets.} We evaluate
  Support2Sim on 50 scenes from four indoor datasets, spanning cluttered
  tabletops and room-scale environments. We also demonstrate how
  reconstructed scenes can be used to train robot policies in simulation
  (Sec.~\ref{sec:policy-learning}).
\end{enumerate}

After 10 seconds of simulation, Support2Sim achieves $51.7\%$ source-mask
intersection over union (IoU), realizes 325 of 373 intended supports, and
has a mean object drift of $45.3$\,mm (Table~\ref{tab:benchmark50}).
Mean IoU decreases by less than one percentage point (from $t_0$ to $t_{10}$)
from $52.6\%$ at release.
Mean generation time ranges from $6.0$ to
$18.8$ minutes by dataset after shared preparation of RGB images,
depth inferred by DA3, and object meshes (Table~\ref{tab:benchmark50}).

\section{Related Work}
\label{sec:related-work}

\subsection{3D Scene Reconstruction}
Neural radiance fields and 3D Gaussian splatting have emerged as leading
representations for multi-view 3D scene reconstruction, capturing detailed
scene appearance for photorealistic novel-view synthesis and robotic
perception and mapping~\cite{mildenhall2020nerf,kerbl2023gaussians,chopra2024agrinerf,chopra2025splatblox}.
Diffusion-based scene generators include Bolt3D, which uses latent diffusion
to generate 3D Gaussians from one or more images~\cite{szymanowicz2025bolt3d},
and Lyra, which learns a 3D Gaussian decoder through video diffusion
self-distillation~\cite{bahmani2025lyra}. MetaScenes converts
3D scans into editable indoor replicas~\cite{yu2025metascenes}, while asset-centric mapping builds
RGB-D maps with explicit object meshes and poses~\cite{hsu2026assetcentric}.

Among image-to-3D models, TRELLIS generates assets with rectified flow~\cite{xiang2025trellis}, SAM~3D
reconstructs masked objects and poses with latent flow matching~\cite{chen2026sam3d},
and MIDI and SceneGen generate compositional scenes using multi-instance
diffusion and feed-forward reconstruction, respectively~\cite{huang2025midi,meng2026scenegen}. InstaScene combines Gaussian-splatting reconstruction with generative object
completion~\cite{yang2025instascene}, while DP-Recon optimizes neural fields
with diffusion priors~\cite{ni2025dprecon}. RecGen uses rectified flow to
reconstruct geometry and poses from sparse RGB-D views~\cite{zadaianchuk2026recgen},
and GATOR combines flow-based reconstruction with agentic mesh and texture
refinement~\cite{wu2026gator}. PhysGS augments reconstructions
with dense physical property estimates~\cite{chopra2026physgs}.
Visual reconstruction alone does not ensure simulation readiness:
our work, Support2Sim, assembles reconstructed assets in a physics simulator to
preserve the source layout, realize inferred support relationships, and
maintain stability under gravity.

\subsection{Real2Sim}
Early work inferred support relations from RGB-D images and used physical
stability to group scene primitives~\cite{silberman2012support,guo2013support,zheng2013beyond}.
CAST imposes signed-distance constraints, PhysOmni aligns objects to a
shared ground plane, and $\phi$-Scene settles objects in support
order~\cite{yao2025cast,zhang2026physomni,li2026phiscene}.
Picasso uses rejection sampling over pose and scale in contact-graph
order, selecting candidates by depth fit under geometric contact and
non-penetration constraints; simulation evaluates the resulting
scenes~\cite{yu2026picasso}.
Physics-Constrained Real2Sim jointly refines pose, mass, friction, and
center of mass through differentiable simulation~\cite{xiang2026physicsreal2sim}.
SPARCS jointly optimizes shape and pose under differentiable force-balance
constraints~\cite{huang2026sparcs}. 

URDFormer, DRAWER, and URDF-Anything+ recover articulated scenes or assets
from images or video~\cite{chen2024urdformer,xia2025drawer,wu2026urdfanything}.
Deformable-body simulation and learned world models capture deformable
dynamics~\cite{xie2024physgaussian,yang2025physworld,zhang2025softbodyeval}.
Vysics and Scalable Real2Sim use robot interaction to infer geometry and
dynamics~\cite{bianchini2025vysics,pfaff2025scalable}.

Reconstructed scenes provide demonstrations for manipulation and
loco-manipulation~\cite{tornevillasevil2024rialto,xiang2026functionpreserving,wang2026vlk}.
MolmoB0T generates demonstrations in MolmoSpaces~\cite{kim2026molmospaces,deshpande2026molmobot},
and SCORE uses a policy pretrained on real data to guide reinforcement
learning in simulation~\cite{yu2026score}. Simulation also enables repeatable policy
evaluation under controlled perturbations~\cite{li2025simpler,jain2025polaris,yang2026robolab}.
GSWorld and DISCOVERSE pair Gaussian rendering with
physics~\cite{jiang2025gsworld,jia2025discoverse}. Indoor scenes can also
be generated from text prompts~\cite{pfaff2026scenesmith}. SimFoundry and
Agentic Real2Sim build interactive environments from
video~\cite{ranawaka2026simfoundry,chen2026agenticreal2sim}. In contrast, our work, Support2Sim, ranks poses from paired surface contacts by
source-mask IoU and uses short rollouts to guide acceptance and compute
allocation. This gradient-free search holds geometry and physical
parameters fixed while pursuing our
\hyperref[obj:system-design]{System Design Objective} within fixed
wall-clock budgets.
\section{Method}
\label{sec:method}

\begin{figure*}[t]
  \centering
  \includegraphics[width=.95\textwidth]{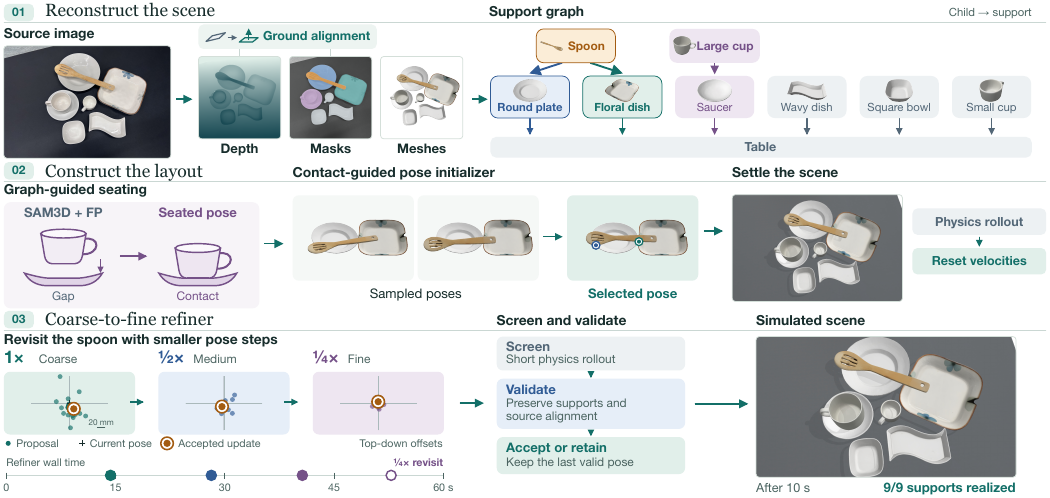}
   \caption{\textbf{Support2Sim architecture.}
  (1) Mesh reconstruction, gravity alignment, and support-graph inference
  from one image. (2) Graph-guided seating and pose initialization, followed
  by settling and velocity reset. (3) Short rollouts guide refinement at
  finer scales using source-mask IoU, support loading, motion, and
  penetration. The final render shows the scene after 10 seconds.}
  \label{fig:architecture}
  \vspace{-.7cm}
\end{figure*}

Given a single RGB image, Support2Sim reconstructs a simulation-ready scene
with visual meshes, collision geometry, object poses, and physical parameters
(Fig.~\ref{fig:architecture}). We infer a graph of intended support
relationships (Sec.~\ref{sec:scene-evidence}) and estimate a ground plane
from depth to establish a world coordinate frame with a gravity-aligned up
direction. We use the graph to seat objects (Sec.~\ref{sec:ground-seating})
and constrain pose sampling for objects with multiple supports
(Sec.~\ref{sec:contact-initialization}). We allocate
separate per-scene wall-clock search budgets $B_{\mathrm{init}}$ and
$B_{\mathrm{ref}}$ to the initializer and refiner, respectively. After
settling the scene, the refiner (Sec.~\ref{sec:coarse-to-fine-refinement})
prioritizes objects and evaluates pose updates in short physics rollouts.

\subsection{Scene evidence and support graph}
\label{sec:scene-evidence}

\noindent\textbf{Scene evidence.}
From the source RGB image, a vision--language model (VLM) identifies the
rigid objects to reconstruct, including their supporting objects, and
predicts a label and bounding box for each instance. We use these labels
and boxes to prompt SAM~3~\cite{carion2026sam3} for instance masks $M_i$.
Depth Anything~3 (DA3)~\cite{lin2026depthanything3} estimates the depth map
$D$ and camera intrinsics $K$. SAM~3D Objects~\cite{chen2026sam3d} reconstructs a
textured mesh from the source image and mask for each instance. The masks
also provide a reference for comparing rendered silhouettes during pose
search. We use VLM-estimated object masses for simulation.
Section~\ref{sec:implementation-details} explains how we estimate each
object's center of mass and set the other simulation parameters.

\noindent\textbf{Support graph.}
We show the VLM the original image alongside a copy annotated with mask
contours and object IDs. It then infers direct relationships between those
instances. For layout construction, we use the inferred resting relations
to form a directed acyclic graph (DAG) $G=(V,E)$, where an edge $i \to j$
means that object $i$ is inferred to rest on object $j$.The target object $i$ and its direct supports
$\mathcal{S}_i$ form a \emph{support group}. For example, the spoon in
Fig.~\ref{fig:architecture} forms a group with the round plate and floral dish.
The support graph is the central data structure linking perception to
simulation: it sets seating order, guides pose initialization, and defines
refinement groups.

\subsection{Ground alignment and graph-guided seating}
\label{sec:ground-seating}

\noindent\textbf{Ground alignment and initial poses.}
We back-project the depth map $D$ using camera intrinsics $K$, both estimated
by DA3, and fit a plane to candidate ground or support points with RANSAC.
The plane normal defines the up direction of a gravity-aligned world
coordinate frame. A visible horizontal support such as a tabletop can
provide the plane when the floor is out of view.
FoundationPose~\cite{wen2024foundationpose} uses the source
RGB image, depth $D$, instance mask $M_i$, and intrinsics $K$ to
register each reconstructed mesh to the source camera. We express each
estimated pose as a rigid transform $T_i \in \mathrm{SE}(3)$ in this world
frame.

\begin{figure*}[t]
  \centering
  \includegraphics[width=.94\textwidth]{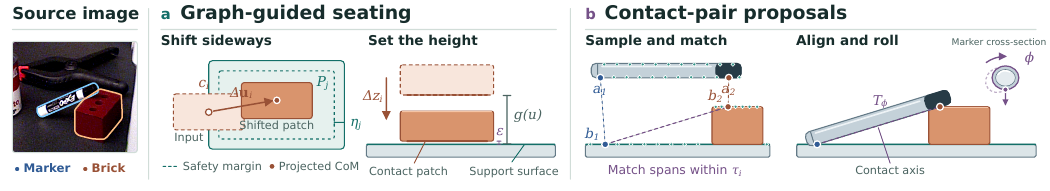}
  \caption{\textbf{Geometry of seating and contact-pair initialization.}
  (a) A horizontal shift places the projected center of mass inside the
  support margin; a vertical shift establishes a small clearance.
  (b) Sampled surface-point pairs with compatible spans define a candidate
  pose. The end view shows the marker cross-section before and after the
  roll determined analytically from surface-normal alignment.}
  \label{fig:contact-geometry}
  \vspace{-.8cm}
\end{figure*}

\noindent\textbf{Graph-guided seating.}
We traverse the support graph in reverse topological order (since edges
point from children to their supporting parents; Fig.~\ref{fig:architecture}),
seating each object after its supports are in place. 
For an object with a
single support $j$, we extract \emph{lower contact patches} from the
downward-facing regions near the object's base. The \emph{support footprint}
$P_j$ is the top-down outline of the support's upward-facing surface
(Fig.~\ref{fig:contact-geometry}a). Let $\mathbf c_i$ be the center of mass
projected onto the world $xy$ plane. We generate a finite set $\mathcal{D}_i$
of translations $\mathbf v=(\Delta x,\Delta y)$ from the projected center
of mass and the contact patch centers. For each candidate, the shifted
center of mass $\mathbf c_i+\mathbf v$ lies inside $P_j$, at least $\eta_j$
from its boundary. The safety margin is
$\eta_j=\min(10\,\mathrm{mm},0.8r_j)$, where $r_j$ is the radius of the
largest circle that fits inside the footprint. For a shift $\mathbf v$, let $\rho(\mathbf v)$
be the fraction of the projected patch area that overlaps $P_j$, and
$\rho_{\max}=\max_{\mathbf v\in\mathcal{D}_i}\rho(\mathbf v)$. We choose the
smallest shift that retains at least $90\%$ of the best candidate overlap,
\vspace{-.2cm}
\begin{equation}
  \Delta\mathbf u_i =
  \operatorname*{argmin}_{\substack{\mathbf v\in\mathcal{D}_i\\
                     \rho(\mathbf v)\ge 0.9\rho_{\max}}} \|\mathbf v\|_2.
  \label{eq:seating-shift}
\end{equation}
The $10\%$ tolerance favors small pose corrections with nearly the same
contact coverage, helping preserve alignment with the source image.
After shifting, we sample horizontal locations $\mathcal{U}_i$ where the
contact patches overlie the support, defining the signed gap $g(\mathbf u)$
as the object's underside height minus the support surface height.
The vertical shift $\Delta z_i=\varepsilon-\min_{\mathbf u\in\mathcal{U}_i}g(\mathbf u)$
raises or lowers the object to leave a minimum sampled gap of
$\varepsilon=0.1\,\mathrm{mm}$ before gravity establishes contact during settling.
We check the seated pose for penetration using the collision geometry.
Objects with multiple supports pass to the contact-guided pose initializer.

\subsection{Contact-guided pose initialization}
\label{sec:contact-initialization}

\noindent\textbf{Contact-pair proposals.}
For an object $i$ with two intended supports, we retain up to 96 points on
its external collision surface and 64 upward-facing points on each support.
We cast rays through the object's source-mask pixels and a narrow band of
neighboring pixels to find support points, then distribute samples using
farthest-point sampling.
We take one point from each support and offset each
slightly along its surface normal for clearance, giving target points
$\mathbf b_1,\mathbf b_2$ in the scene frame. Since a rigid motion preserves
distances, we match them to object points $\mathbf a_1,\mathbf a_2$ in its
local frame, separated by at least $0.15L_i$ and satisfying
$\bigl|\|\mathbf a_2-\mathbf a_1\|_2-\|\mathbf b_2-\mathbf b_1\|_2\bigr|\le\tau_i$.
We use $\tau_i=\min(0.01L_i,1\,\mathrm{mm})$, where $L_i$ is the reconstructed
mesh's extent along its dominant principal axis
(Fig.~\ref{fig:contact-geometry}b).


Starting from the input rotation, we choose $R_0$ to align the direction of
the object pair with the direction $\hat{\mathbf d}_b$ from $\mathbf b_1$
to $\mathbf b_2$. The pair midpoints $\bar{\mathbf a}$ and $\bar{\mathbf b}$
determine the translation. A rotation $\phi$ about the contact axis gives
the candidate pose $T_\phi=(R_\phi,\mathbf t_\phi)$,
\vspace{-.2cm}
\begin{equation}
  R_\phi = \operatorname{Rot}(\hat{\mathbf d}_b,\phi)R_0,\quad \mathbf t_\phi = \bar{\mathbf b}-R_\phi\bar{\mathbf a}.
  \label{eq:contact-pair-pose}
  \vspace{-.2cm}
\end{equation}
We compute $\phi$ analytically to maximize opposing surface-normal
alignment at both proposed contacts.
Repeating this construction generates up to 6,000 candidate poses per scene
within $B_{\mathrm{init}}$; the following stage ranks these poses for
detailed geometric screening.

\noindent\textbf{Geometric screening and selection.}
Consistent with our \hyperref[obj:system-design]{System Design Objective},
we rank candidate poses before costly collision checks and rendering,
directing $B_{\mathrm{init}}$ toward promising candidates.
The preliminary score prioritizes source-mask IoU, approximated using the
convex hull of projected mesh points, with smaller penalties for
surface-normal mismatch and changes from the input pose.

Within each support group, we check candidates in ranked order for pose
bounds, scene interpenetration, and the gap and upward normal at both
supports. Those passing pose and collision checks are rendered within the
group's share of $B_{\mathrm{init}}$. Let $\mathcal{F}_i$ contain the rendered pose candidates
that also pass both support checks, and let $\widehat{M}_i(T_\phi)$ be the
rendered source-view silhouette. We select
\vspace{-.3cm}
\begin{equation}
  T_i^\star = \operatorname*{argmax}_{T_\phi\in\mathcal{F}_i}
  \operatorname{IoU}\bigl(\widehat{M}_i(T_\phi), M_i\bigr).
  \label{eq:initializer-selection}
  \vspace{-.4cm}
\end{equation}
If no candidate pose passes the pose bounds and collision checks,
we keep the current pose. We settle the assembled layout under gravity to
let contacts form and initial motion subside, then reset linear and angular
velocities so subsequent rollouts assess stability from rest.

\subsection{Coarse-to-fine refinement}
\label{sec:coarse-to-fine-refinement}

Contacts that satisfy geometric checks can still separate under gravity.
The coarse-to-fine refiner evaluates dynamic stability in physics rollouts
by measuring contact persistence and object motion, then searches for pose
updates that preserve intended supports and reduce drift while maintaining
source-image agreement.


\begin{figure}[t]
  \centering
  \includegraphics[width=.95\columnwidth]{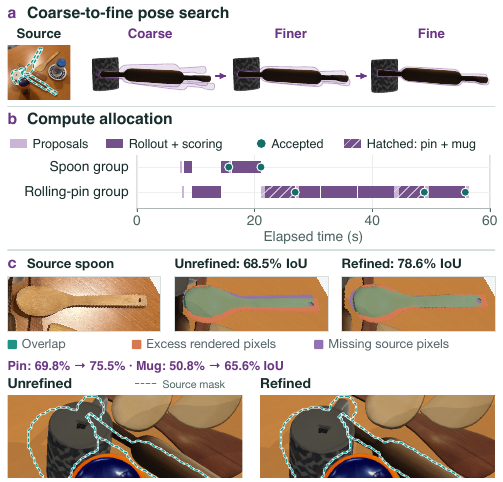}
  \caption{\textbf{Coarse-to-fine refinement in one scene.}
  (a) Source image and ghosted rolling-pin poses at progressively finer
  scales. (b) Recorded 60-second allocation by support group: light and
  dark bars show proposal work and rollout scoring; dots mark accepted
  updates. Hatched intervals move the pin and mug jointly within the
  rolling-pin group. (c) Source and 10-second renders. Spoon overlays show
  agreement (teal), excess rendered pixels (orange), and missing source
  pixels (purple).
  Full-object IoUs are shown;
  mean IoU over four refinement objects improves from $64.5\%$ to $72.2\%$,
  retaining all 10 intended supports.}
  \label{fig:refinement-loop}
  \vspace{-.8cm}
\end{figure}

\noindent\textbf{Select a support group.}
Starting from the settled layout, we allocate the separate refinement
budget $B_{\mathrm{ref}}$ across support groups. Each \emph{visit} allocates
time to one support group for pose generation and geometric screening
(light bars), then physics rollouts and scoring (dark bars in
Fig.~\ref{fig:refinement-loop}b).

We prioritize \emph{support groups} using the latest rollout of the current layout.
Over its final second, $\ell_{ij}$ is the fraction of samples in which
the pair $(i,j)$ has force-bearing contact. We set
$f_i=1$ (unintended contacts) if, during that second, $i$ makes force-bearing
contact with or comes too close to an object that should stay separate,
and $f_i=0$ otherwise. We measure
motion $m_i$ as the largest distance between any two sampled positions of
$i$ in that second. Let $I_i$ be the target's source-mask IoU at the end
of the rollout. We define the support deficit
$d_i=\sum_{j\in\mathcal{S}_i}(1-\ell_{ij})$ and assign each \emph{support group}
the priority tuple
\vspace{-.25cm}
\begin{equation}
  \mathbf p_i = \bigl(f_i,d_i,1-I_i,m_i\bigr).
  \label{eq:refiner-priority}
  \vspace{-.3cm}
\end{equation}
We rank support groups by $\mathbf p_i$ from left to right, preferring larger
entries. After one pass, we revisit the highest-ranked group with search
options remaining and update priorities from accepted rollouts. If no
target has multiple supports, we consider single-support objects that
violate contact, motion, or penetration limits.

\noindent\textbf{Propose pose updates.}
We first sample translations and rotations of target $i$ alone. The three
search scales use local translation ranges of $\pm15\%$, $\pm7.5\%$, and
$\pm3.75\%$ of the object's largest principal-axis extent along each sampled
direction, paired with rotation radii of $5^\circ$, $2.5^\circ$, and
$1.25^\circ$, respectively. We reject candidates that lose established
supports, violate separation requirements for excluded object pairs,
exceed penetration or motion thresholds during rollout, or reduce
source-mask IoU in the surrounding scene beyond tolerance. Accepted updates
halve these sampling ranges down to the finest scale. 
If target-only
proposals fail, we retain the scale and consider related movable objects,
individually or jointly with the target. This enables coupled corrections
such as moving a rolling pin with its supporting mug
(Fig.~\ref{fig:refinement-loop}b).

\noindent\textbf{Roll out, accept, and narrow the search ranges.}
We run short physics rollouts for geometrically shortlisted candidates
(dark bars in Fig.~\ref{fig:refinement-loop}b), then render valid terminal poses to compute
\vspace{-.2cm}
\begin{equation}
  \mathcal{L} = w_I(1-\bar I) + w_S\mathcal{L}_S
  + w_M\mathcal{L}_M + w_P\mathcal{L}_P ,
  \label{eq:refiner-loss}
  \vspace{-.2cm}
\end{equation}
where $\bar I$ is the mean source-mask IoU over the refinement scope and
$\mathcal{L}_S$ penalizes unloaded intended supports over the final second.
The motion loss is
$\mathcal{L}_M=\max_{i\in\mathcal{R}}\max\{m_i/(2\,\mathrm{mm}),\,\theta_i/(3^\circ)\}$,
where $\mathcal{R}$ contains the movable objects in the refinement scope;
$m_i$ and $\theta_i$ are the largest position and orientation differences
between sampled poses of object $i$ in the final second. We normalize the
maximum terminal penetration by $0.1\,\mathrm{mm}$ to obtain $\mathcal{L}_P$.
We use $(w_I,w_S,w_M,w_P)=(1,2,0.05,0.02)$, emphasizing support realization
and source-mask IoU while penalizing motion and penetration.

We accept feasible updates that recover an intended support or lower
$\mathcal{L}$ beyond tolerance while preserving supports.
After longer-rollout validation, we release the settled poses from rest;
if validation fails, we retain the initial settled layout
(Fig.~\ref{fig:refinement-loop}c).

\section{Results}
\label{sec:results}

\subsection{Implementation details}
\label{sec:implementation-details}
We prepare object inventories, masks, inferred depth, and meshes once per
scene for reuse across methods. GPT-6.1 Sol with high reasoning effort
generates the inventories and support graphs. Local stages of the system run on a
workstation with an AMD Ryzen Threadripper 3960X CPU and an NVIDIA RTX 6000
Ada GPU.

We decompose visual meshes into convex collision parts with V-HACD and
simulate rigid bodies in PyBullet at 240\,Hz with gravitational
acceleration $g=9.8\,\mathrm{m/s^2}$. For movable bodies, we use lateral
friction of $1.0$ and zero restitution, rolling friction, and spinning
friction. We compute
the center of mass as the centroid of the volume enclosed by the visual
mesh under uniform density, using the collision geometry when the visual
mesh is open or invalid.

We set separate per-scene wall-clock search budgets of
$B_{\mathrm{init}}=60\,\mathrm{s}$ for the initializer and
$B_{\mathrm{ref}}=60\,\mathrm{s}$ for the refiner.
The refiner tests proposals
with 2-second rollouts and validates the selected layout with 10-second
rollouts. Validation limits each movable body's peak displacement and
rotation to 10\,mm and $10^\circ$, respectively.

\subsection{Reconstruction benchmark}
\label{sec:reconstruction-benchmark}

\noindent\textbf{Datasets.}
We evaluate 50 scenes from four datasets: 25 from
MessyKitchens~\cite{ansari2026messykitchens}, 10 from
YCB-Video~\cite{xiang2018posecnn}, 10 from Picasso~\cite{yu2026picasso},
and five from HyperSim~\cite{roberts2021hypersim}. The scenes span cluttered
tabletops and furnished indoor rooms, including stacked objects and objects
with multiple supports.

\noindent\textbf{Metrics.}
\label{sec:evaluation-metrics}
For each scene, methods share one fixed RGB view and the same object meshes.
We evaluate reconstructions using a common 10-second physics rollout and
score them against the same reference camera, source masks, and intended
support edges.
\begin{itemize}[nosep]
  \item \textbf{Source-mask IoU.} We compare the rendered silhouette $R$
  with the source mask $M$ using $|R\cap M|/|R\cup M|$ at release ($t_0$)
  and after 10 seconds ($t_{10}$). We average over objects, then over scenes.
  \item \textbf{Intended supports.} We pool realized and eligible edges
  from a shared, frozen inferred support graph. 
  We use a normal contact-force threshold of \(10^{-6}\) N to assess support realization.
  Edges involving missing objects count as unrealized.
  \item \textbf{Drift.} We average the Euclidean displacement
  $\|\mathbf{x}_i(t_{10})-\mathbf{x}_i(t_0)\|_2$ first over objects, then
  over scenes, where $\mathbf{x}_i(t)$ denotes object $i$'s position in
  millimeters.
  \item \textbf{Severe failures (\textcolor[HTML]{B23A48}{$\times$}).}
  We count a scene once if any object moves more than 1\,m from release
  (2\,m for HyperSim). We use the higher HyperSim cutoff because its scenes
  contain roughly twice as many objects as the other datasets
  (17.2 versus 7.4 objects per scene on average).
  \item \textbf{Generation time.} Wall-clock mean $\pm$ standard deviation
  after shared RGB, DA3 depth, and mesh preparation; omitted for the
  preprocessing baselines SAM3D and SAM3D + FP.
\end{itemize}

\begin{table*}[!t]
  \centering
  \caption{\textbf{50-scene evaluation across four datasets.}
  \textcolor[HTML]{165D55}{\textbf{Bold teal}} marks the best result for each metric within each dataset.
  The red \textcolor[HTML]{B23A48}{$\times$} denotes the number of severe scene failures.}
  \vspace{-.3cm}
  \label{tab:benchmark50}
  \small
  \begingroup
  \definecolor{cmpMK}{HTML}{E3F0EB}
  \definecolor{cmpYCB}{HTML}{E8EEF8}
  \definecolor{cmpPicasso}{HTML}{F0E9F5}
  \definecolor{cmpHyperSim}{HTML}{F5EEDC}
  \definecolor{cmpAccent}{HTML}{165D55}
  \definecolor{cmpMuted}{HTML}{69747A}
  \definecolor{cmpFailure}{HTML}{B23A48}
  \newcommand{\cmpBest}[1]{%
    \textcolor{cmpAccent}{%
      \ifdefined\XeTeXversion
        \fontfamily{lmr}\fontseries{bx}\selectfont #1%
      \else
        \textbf{#1}%
      \fi}}
  \newcommand{\cmpHeader}[3]{%
    \begingroup
    \setlength{\fboxsep}{2pt}%
    \colorbox{#1}{\makebox[\dimexpr\linewidth-2\fboxsep\relax][l]{%
      \strut\textbf{#2}\hfill\textcolor{cmpMuted}{#3 scenes}}}%
    \endgroup\par\vspace{1pt}%
  }
  \newcommand{\cmpNA}{\textcolor{cmpMuted}{\textemdash}}
  \newcommand{\cmpTime}[2]{\mbox{#1\,$\pm$\,#2}}
  \newcommand{\cmpColumns}{%
    \raisebox{-7pt}[0pt][0pt]{Method} & \multicolumn{2}{c}{IoU (\%) $\uparrow$} & \multicolumn{1}{c}{Intended} & \raisebox{-7pt}[0pt][0pt]{Drift (mm) $\downarrow$} & \raisebox{-7pt}[0pt][0pt]{\textcolor{cmpFailure}{$\times$}\,$\downarrow$} & \raisebox{-7pt}[0pt][0pt]{Time (min) $\downarrow$} \\
    \cmidrule(lr){2-3}
    & $t_0$ & $t_{10}$ & \multicolumn{1}{c}{supports $\uparrow$} & & & \\
    \midrule
  }
  \setlength{\tabcolsep}{1.2pt}
  \renewcommand{\arraystretch}{1.0}
  \resizebox{.95\textwidth}{!}{%
  \begin{minipage}{\textwidth}
  \begin{minipage}[t]{.49\textwidth}
    \cmpHeader{cmpMK}{MessyKitchens}{25}
    \begin{tabular*}{\linewidth}{@{\extracolsep{\fill}}lrrrrrr@{}}
      \cmpColumns
      SAM3D           & 46.3 & 22.9 & 70/176 & 928.18 & 24 & \cmpNA \\
      SAM3D + FP      & \cmpBest{76.8} & 29.4 & 97/176 & 633.39 & 20 & \cmpNA \\
      SimFoundry      & 42.7 & 40.7 & 157/176 & 23.26 & \cmpBest{0} & \cmpTime{18.9}{3.7} \\
      PCR             & 38.4 & 29.2 & 89/176 & 171.25 & 3 & \cmpTime{24.5}{15.6} \\
      Picasso         & 44.3 & 16.4 & 122/176 & 872.84 & 22 & \cmpBest{\cmpTime{1.0}{0.3}} \\
      \midrule
      Support2Sim     & 61.4 & \cmpBest{61.4} & \cmpBest{162/176} & \cmpBest{4.37} & \cmpBest{0} & \cmpTime{10.1}{1.9} \\
      \bottomrule
    \end{tabular*}
  \end{minipage}\hfill
  \begin{minipage}[t]{.49\textwidth}
    \cmpHeader{cmpYCB}{YCB-Video}{10}
    \begin{tabular*}{\linewidth}{@{\extracolsep{\fill}}lrrrrrr@{}}
      \cmpColumns
      SAM3D           & 50.9 & 11.8 & 14/52 & 1,531.58 & 9 & \cmpNA \\
      SAM3D + FP      & \cmpBest{74.7} & 6.3 & 13/52 & 1,135.32 & 10 & \cmpNA \\
      SimFoundry      & 16.9 & 14.2 & 44/52 & 10.99 & \cmpBest{0} & \cmpTime{16.2}{12.2} \\
      PCR             & 11.3 & 11.6 & 15/52 & 378.51 & 7 & \cmpTime{9.6}{7.3} \\
      Picasso         & 42.2 & 14.7 & 32/52 & 1,258.60 & 10 & \cmpBest{\cmpTime{0.5}{0.3}} \\
      \midrule
      Support2Sim     & 57.7 & \cmpBest{57.7} & \cmpBest{50/52} & \cmpBest{0.06} & \cmpBest{0} & \cmpTime{6.0}{0.7} \\
      \bottomrule
    \end{tabular*}
  \end{minipage}

  \vspace{4pt}

  \begin{minipage}[t]{.49\textwidth}
    \cmpHeader{cmpPicasso}{Picasso}{10}
    \begin{tabular*}{\linewidth}{@{\extracolsep{\fill}}lrrrrrr@{}}
      \cmpColumns
      SAM3D           & 40.3 & 11.0 & 27/59 & 1,091.81 & 7 & \cmpNA \\
      SAM3D + FP      & \cmpBest{56.6} & 15.4 & 32/59 & 945.40 & 7 & \cmpNA \\
      SimFoundry      & 28.3 & 28.4 & 39/59 & \cmpBest{15.58} & \cmpBest{0} & \cmpTime{11.0}{3.4} \\
      PCR             & 31.2 & 14.5 & 30/59 & 538.66 & 5 & \cmpTime{20.4}{12.3} \\
      Picasso         & 45.6 & 13.2 & 32/59 & 14,637.37 & 8 & \cmpBest{\cmpTime{0.6}{0.3}} \\
      \midrule
      Support2Sim     & 36.1 & \cmpBest{32.6} & \cmpBest{44/59} & 127.67 & 1 & \cmpTime{6.5}{0.6} \\
      \bottomrule
    \end{tabular*}
  \end{minipage}\hfill
  \begin{minipage}[t]{.49\textwidth}
    \cmpHeader{cmpHyperSim}{HyperSim}{5}
    \begin{tabular*}{\linewidth}{@{\extracolsep{\fill}}lrrrrrr@{}}
      \cmpColumns
      SAM3D           & 50.0 & 19.2 & 46/86 & 4,286.91 & 5 & \cmpNA \\
      SAM3D + FP      & \cmpBest{55.2} & 18.4 & 40/86 & 3,890.99 & 5 & \cmpNA \\
      SimFoundry      & 11.9 & 14.2 & 33/86 & 626.57 & \cmpBest{1} & \cmpTime{48.9}{28.0} \\
      PCR             & 50.5 & 19.4 & 40/86 & 4,132.69 & 5 & \cmpTime{22.1}{12.3} \\
      Picasso         & 24.7 & 3.4 & 26/86 & 16,922.44 & 5 & \cmpBest{\cmpTime{1.5}{0.9}} \\
      \midrule
      Support2Sim     & 31.1 & \cmpBest{29.7} & \cmpBest{69/86} & \cmpBest{175.51} & \cmpBest{1} & \cmpTime{18.8}{7.5} \\
      \bottomrule
    \end{tabular*}
  \end{minipage}

  \end{minipage}%
  }
  \endgroup
  \vspace{-.4cm}
\end{table*}

\begin{figure*}[!t]
  \centering
  \includegraphics[width=.94\textwidth]
    {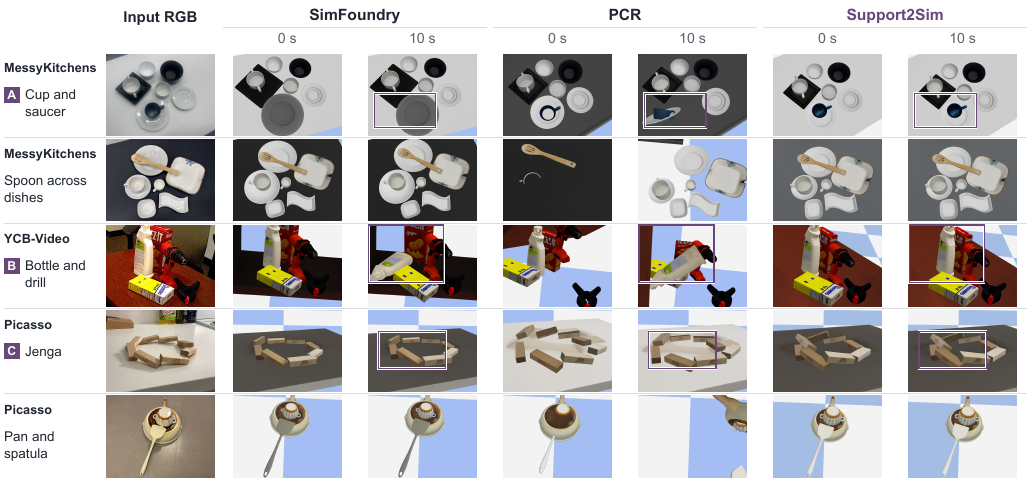}
  \caption{\textbf{Qualitative comparison on tabletop scenes.}
  Matched source-camera views at release and after 10 seconds.
  Support2Sim has higher final IoU than both shown baselines in the cup
  (A), spoon, bottle (B), and pan scenes. In the bottle scene, final IoU is
  $75.4\%$, versus $33.2\%$ for SimFoundry and $25.9\%$ for PCR.
  Jenga (C) highlights intended-support realization: $14/14$ for
  Support2Sim, versus $9/14$ and $10/14$, respectively.}
  \label{fig:qualitative-tabletop}
  \vspace{-.7cm}
\end{figure*}

\begin{table}[!t]
  \centering
  \caption{\textbf{Cumulative ablation} on 16 contact-rich scenes with 10-second rollouts.}
  \vspace{-.2cm}
  \label{tab:ablation}
  \small
  \setlength{\tabcolsep}{2.5pt}
  \definecolor{ablationAccent}{HTML}{165D55}
  \newcommand{\ablationBest}[1]{\textcolor{ablationAccent}{\textbf{#1}}}
  \resizebox{.95\columnwidth}{!}{%
  \begin{tabular*}{\columnwidth}{@{\extracolsep{\fill}}lrrr@{}}
    \toprule
    \raisebox{.55em}{System stage} &
    \shortstack[r]{IoU $\uparrow$\\(\%)} &
    \raisebox{.55em}{Supports $\uparrow$} &
    \shortstack[r]{Drift $\downarrow$\\(mm)} \\
    \midrule
    SAM3D + FP & 23.11 & 72/129 & 41.238 \\
    + Graph-guided seating & 50.73 & 116/129 & 0.270 \\
    + Pose initializer & 52.57 & 121/129 & 0.191 \\
    \addlinespace[2pt]
    + Coarse-to-fine refiner & \ablationBest{53.69} &
    \ablationBest{123/129} & \ablationBest{0.172} \\
    \bottomrule
  \end{tabular*}%
  }
  \vspace{-.4cm}
\end{table}


\noindent\textbf{Baselines.}
\label{sec:baseline-comparison}
We compare with five baselines.
\begin{itemize}[nosep]
  \item \textbf{SAM3D.} Uses the reconstructed meshes and initial poses
  estimated by SAM3D.
  \item \textbf{SAM3D + FP.} Refines the SAM3D poses with FoundationPose
  using RGB-D observations.
  \item \textbf{SimFoundry.} Fits object poses using the shared RGB images
  and depth inferred by DA3, then stabilizes the assembled scene with physics.
  \item \textbf{Physics-Constrained Real2Sim (PCR).} Registers meshes to
  depth with ICP, followed by geometric and physics-constrained pose
  optimization.
  \item \textbf{Picasso.} Fits meshes to depth using rejection sampling
  guided by a contact scene graph.
\end{itemize}

\noindent\textbf{Quantitative comparison.}
Across 50 scenes, Support2Sim's mean source-mask IoU decreases from
$52.6\%$ at $t_0$ to $51.7\%$ at $t_{10}$, a decline of less than one
percentage point (Table~\ref{tab:benchmark50}). SAM3D + FoundationPose
falls from $70.2\%$ to $20.9\%$, underscoring the importance of preserving
source-image agreement during simulation.

Support2Sim leads all four datasets in mean final IoU and intended-support
realization, with the highest final IoU on 39 of 50 scenes. It realizes
$87.1\%$ of intended supports (325/373), compared with $73.2\%$ (273/373)
for SimFoundry, and reduces mean drift by $43.1\%$, from $79.6$ to
$45.3$\,mm.
Mean generation times are lower than SimFoundry and PCR across datasets.
Picasso is faster, but has lower final IoU and fewer realized supports.

\noindent\textbf{MessyKitchens.}
Support2Sim preserves mean IoU from $t_0$ to $t_{10}$ at the reported
precision and leads final IoU on 20 of 25 scenes. In the cup-and-dish
scene, SimFoundry has higher final IoU ($77.8\%$ versus $54.5\%$), while
Support2Sim realizes all three intended supports, compared with two.

\noindent\textbf{YCB-Video.}
Mean IoU likewise remains unchanged at the reported precision, and
Support2Sim leads final IoU on eight of ten scenes. On the mug-and-can stack,
SimFoundry has higher final IoU ($30.4\%$ versus $15.7\%$), but realizes
only two of six intended supports, compared with five for Support2Sim.

\noindent\textbf{Picasso.}
Mean IoU drops by $3.5$ percentage points but remains the highest among
methods. Support2Sim leads final IoU on six of ten scenes.
Although Support2Sim has lower drift than SimFoundry in seven of ten scenes,
larger errors in the remaining scenes raise its mean drift.

\noindent\textbf{HyperSim.}
Support2Sim leads final IoU in all five rooms, with a mean IoU drop of
$1.4$ percentage points. In the kitchen it
realizes 17 of 21 supports, while the best baseline realizes eight.
Construction and seating dominate the longer generation time in these rooms.
The dining room exposes residual instability: SimFoundry has much lower
drift ($26.9$ versus $761.0$\,mm), while Support2Sim retains higher final
IoU ($20.4\%$ versus $4.8\%$) and realizes more intended supports
(15/28 versus 6/28).

\begin{figure}[!t]
  \centering
  \includegraphics[width=.95\columnwidth]{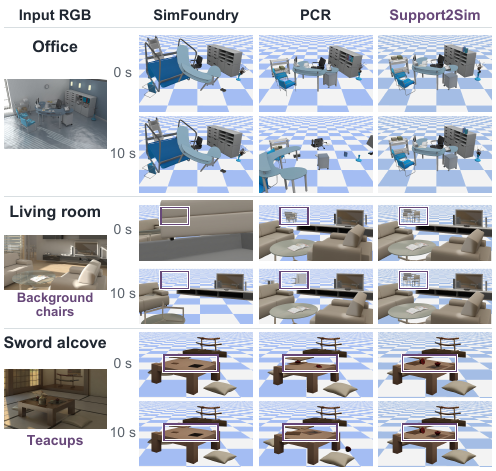}
  \caption{\textbf{Qualitative comparison on HyperSim.}
  Matched source-camera views at release and after 10 seconds. Boxes
  highlight the living-room chairs and the teacups in the sword alcove.
  In PCR, the chairs tip over and a cup falls off the table. Support2Sim
  retains both arrangements.}
  \label{fig:qualitative-hypersim}
  \vspace{-.7cm}
\end{figure}

\noindent\textbf{Qualitative comparison.}
\label{sec:qualitative-comparison}
In MessyKitchens, the cup remains seated on its saucer with Support2Sim,
and the YCB-Video bottle stays upright while it tips over in SimFoundry
and PCR (Fig.~\ref{fig:qualitative-tabletop}A--B). In the first Jenga layout,
SimFoundry has slightly higher final IoU, while Support2Sim preserves all
14 intended supports, including block-to-block contacts. SimFoundry
and PCR realize 9 and 10, respectively (Fig.~\ref{fig:qualitative-tabletop}C).
In the HyperSim living room, the background chairs tip over in PCR and
IoU drops from $56.3\%$ to $37.1\%$. With Support2Sim, the chairs remain
upright and IoU stays near $42\%$ (Fig.~\ref{fig:qualitative-hypersim}).

\raggedbottom

\begin{figure}[tb]
  \centering
  \includegraphics[width=.95\columnwidth]{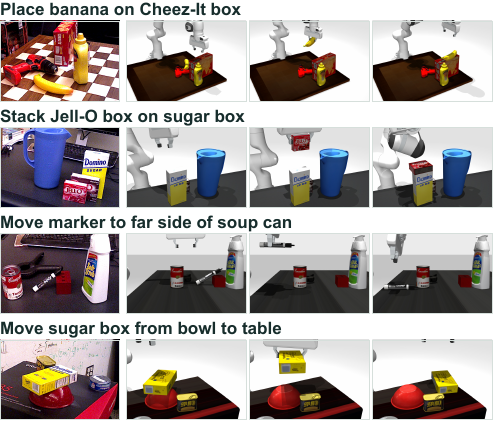}
  \caption{\textbf{Policy rollouts in reconstructed scenes.}
  Each row shows the source photo, initial simulation state, grasp, and
  final placement.}
  \label{fig:policy-learning}
  \vspace{-.7cm}
\end{figure}


\noindent\textbf{Ablation.}
\label{sec:ablation}
We evaluate 16 contact-rich scenes (from MessyKitchens, YCB-Video, and Picasso)
from the 50-scene benchmark, with 129 intended support edges
across cups on saucers, utensils spanning dishes, and block stacks.
The ablation evaluates graph-guided seating (Sec.~\ref{sec:ground-seating}),
the contact-guided pose initializer (Sec.~\ref{sec:contact-initialization}),
and the coarse-to-fine refiner (Sec.~\ref{sec:coarse-to-fine-refinement}).
Graph-guided seating corrects lateral offsets and
contact gaps before release, yielding the largest gain: 44 additional
realized supports and a $27.6$\,pp increase in IoU
(Table~\ref{tab:ablation}). The initializer fits objects across two supports
and ranks feasible poses by source-mask IoU, adding five supports and
$1.8$ IoU points. Rollout-guided refinement adds two more supports and
$1.1$ IoU points, reducing mean drift from $0.191$ to $0.172$\,mm.

\subsection{Applications: Policy learning in simulation}
\label{sec:policy-learning}
To demonstrate Support2Sim's application to robot learning, we train
state-based Franka Panda policies in MuJoCo Playground~\cite{todorov2012mujoco,zakka2025mujocoplayground}
using behavior cloning with
DART noise injection (BC + DART)~\cite{laskey2017dart} and 3,000
scripted-expert demonstrations per task. We add a Franka Panda, position
the four reconstructed YCB-Video scenes within its workspace, and define
goals for banana placement, box stacking, marker relocation, and box transfer
(Fig.~\ref{fig:policy-learning}).
Across 128 held-out randomized starts per task, BC + DART averages
$91.0\%$ success, requiring released, stable placements without disturbing
surrounding objects.

\section{Conclusion, Limitations and Future Work}
\label{sec:conclusion}

We presented Support2Sim, a Real2Sim system that reconstructs
simulation-ready scenes from a single RGB image. Support graphs,
contact-guided pose initialization, and coarse-to-fine refinement balance
source fidelity, support realization, and stability through bounded pose
searches. Across 50 scenes from four datasets, Support2Sim achieves the
highest mean final IoU and intended-support realization among the evaluated
methods, while mean IoU decreases by less than one percentage point over
10 seconds. We also demonstrate policy learning in four reconstructed
scenes from the YCB-Video dataset.

Our system is highly sensitive to the quality of support graphs inferred
by the VLM and outputs from foundation models such as
SAM3D and FoundationPose.
Further, we only handle rigid objects; future work will extend it to
deformable and articulated objects~\cite{cao2026physxanything,zhang2026physomni,zhang2025softbodyeval}.
We will also pursue real2sim2real by
training policies in reconstructed simulations and evaluating them in both
simulation and the real world, with the goal of reducing the sim2real gap
and enabling policy training at scale in simulation.

\section*{LLM Usage Statement}
We used OpenAI Codex to draft and revise text, assist with literature
searches, generate figure code, and carry out multi-step implementation,
debugging, and experiment runs. The authors directed the work and take
responsibility for the submitted results and claims.

{\small
\bibliographystyle{ieeenat_fullname}
\bibliography{references}

@inproceedings{carion2026sam3,
  author = {Carion, Nicolas and Gustafson, Laura and Hu, Yuan-Ting and
            Debnath, Shoubhik and Hu, Ronghang and Suris, Didac and Ryali, Chaitanya and
            Alwala, Kalyan Vasudev and Khedr, Haitham and Huang, Andrew and
            Lei, Jie and Ma, Tengyu and Guo, Baishan and Kalla, Arpit and
            Marks, Markus and Greer, Joseph and Wang, Meng and Sun, Peize and
            R{\"a}dle, Roman and Afouras, Triantafyllos and Mavroudi, Effrosyni and
            Xu, Katherine and Wu, Tsung-Han and Zhou, Yu and Momeni, Liliane and
            Hazra, Rishi and Ding, Shuangrui and Vaze, Sagar and Porcher, Francois and
            Li, Feng and Li, Siyuan and Kamath, Aishwarya and Cheng, Ho Kei and
            Doll{\'a}r, Piotr and Ravi, Nikhila and Saenko, Kate and Zhang, Pengchuan and
            Feichtenhofer, Christoph},
  title = {{SAM 3}: Segment Anything with Concepts},
  booktitle = {International Conference on Learning Representations},
  year = {2026}
}

@inproceedings{lin2026depthanything3,
  author = {Lin, Haotong and Chen, Sili and Liew, Jun Hao and Chen, Donny Y. and
            Li, Zhenyu and Shi, Guang and Feng, Jiashi and Kang, Bingyi},
  title = {{Depth Anything 3}: Recovering the Visual Space from Any Views},
  booktitle = {International Conference on Learning Representations},
  year = {2026}
}

@inproceedings{chen2026sam3d,
  author = {Chen, Xingyu and Chu, Fu-Jen and Gleize, Pierre and Liang, Kevin J and
            Sax, Alexander and Tang, Hao and Wang, Weiyao and Guo, Michelle and
            Hardin, Thibaut and Li, Xiang and Lin, Aohan and Liu, Jia-Wei and Ma, Ziqi and
            Sagar, Anushka and Song, Bowen and Wang, Xiaodong and Yang, Jianing and
            Zhang, Bowen and Doll{\'a}r, Piotr and Gkioxari, Georgia and Feiszli, Matt and
            Malik, Jitendra},
  title = {{SAM 3D}: {3Dfy} Anything in Images},
  booktitle = {Proceedings of the IEEE/CVF Conference on Computer Vision and Pattern Recognition},
  year = {2026},
  pages = {7220--7232}
}

@inproceedings{wen2024foundationpose,
  author = {Wen, Bowen and Yang, Wei and Kautz, Jan and Birchfield, Stan},
  title = {{FoundationPose}: Unified {6D} Pose Estimation and Tracking of Novel Objects},
  booktitle = {Proceedings of the IEEE/CVF Conference on Computer Vision and Pattern Recognition},
  year = {2024},
  pages = {17868--17879}
}

@article{jain2025polaris,
  author = {Jain, Arhan and Zhang, Mingtong and Arora, Kanav and Chen, William and
            Torne, Marcel and Irshad, Muhammad Zubair and Zakharov, Sergey and
            Wang, Yue and Levine, Sergey and Finn, Chelsea and Ma, Wei-Chiu and
            Shah, Dhruv and Gupta, Abhishek and Pertsch, Karl},
  title = {{PolaRiS}: Scalable Real-to-Sim Evaluations for Generalist Robot Policies},
  journal = {arXiv preprint arXiv:2512.16881},
  year = {2025}
}

@article{ranawaka2026simfoundry,
  author = {Ranawaka, Nadun and Wong, Josiah and Pai, Wei-Lin and Chu, Wei-Teng and
            Dai, Tianyuan and Moghani, Masoud and Yin, Hang and Jiang, Yunfan and
            Durbano, Wesley and Huynh, Brandon and Fang, Yu and Xu, Danfei and
            Zhang, Ruohan and Fei-Fei, Li and Fan, Linxi and Wen, Bowen and
            Mandlekar, Ajay and Zhu, Yuke},
  title = {{SimFoundry}: Modular and Automated Scene Generation for Policy Learning and Evaluation},
  journal = {arXiv preprint arXiv:2606.28276},
  year = {2026}
}

@inproceedings{yu2026picasso,
  author = {Yu, Xihang and Talak, Rajat and Shaikewitz, Lorenzo and Carlone, Luca},
  title = {{Picasso}: Holistic Scene Reconstruction with Physics-Constrained Sampling},
  booktitle = {Proceedings of Robotics: Science and Systems},
  year = {2026},
  doi = {10.15607/RSS.2026.XXII.173}
}

@article{li2026phiscene,
  author = {Li, Haodong and Shao, Lulu and Lu, Haolin and Fu, Yu and
            Chen, Yen-Ru and Jain, Seemandhar and Chandraker, Manmohan},
  title = {{$\phi$-Scene}: Physically Grounded Image-to-{3D} Scene Reconstruction},
  journal = {arXiv preprint arXiv:2606.21596},
  year = {2026}
}

@article{xiang2026physicsreal2sim,
  author = {Xiang, Tianyi and Cao, Jiahang and Guo, Sikai and Zhao, Guoyang and
            Luo, Andrew F. and Ma, Jun},
  title = {Real-to-Sim for Highly Cluttered Environments via Physics-Consistent Inter-Object Reasoning},
  journal = {arXiv preprint arXiv:2602.12633},
  year = {2026}
}

@inproceedings{tornevillasevil2024rialto,
  author = {Torne Villasevil, Marcel and Simeonov, Anthony and Li, Zechu and
            Chan, April and Chen, Tao and Gupta, Abhishek and Agrawal, Pulkit},
  title = {Reconciling Reality through Simulation: A Real-to-Sim-to-Real Approach for Robust Manipulation},
  booktitle = {Proceedings of Robotics: Science and Systems},
  year = {2024},
  doi = {10.15607/RSS.2024.XX.015}
}

@inproceedings{nasiriany2026robocasa365,
  author = {Nasiriany, Soroush and Nasiriany, Sepehr and Maddukuri, Abhiram and Zhu, Yuke},
  title = {{RoboCasa365}: A Large-Scale Simulation Framework for Training and Benchmarking Generalist Robots},
  booktitle = {International Conference on Learning Representations},
  year = {2026}
}

@inproceedings{li2025simpler,
  author = {Li, Xuanlin and Hsu, Kyle and Gu, Jiayuan and Mees, Oier and
            Pertsch, Karl and Walke, Homer Rich and Fu, Chuyuan and
            Lunawat, Ishikaa and Sieh, Isabel and Kirmani, Sean and Levine, Sergey and
            Wu, Jiajun and Finn, Chelsea and Su, Hao and Vuong, Quan and Xiao, Ted},
  title = {Evaluating Real-World Robot Manipulation Policies in Simulation},
  booktitle = {Proceedings of the 8th Conference on Robot Learning},
  series = {Proceedings of Machine Learning Research},
  volume = {270},
  pages = {3705--3728},
  year = {2025}
}

@article{liu2026worldgrounding,
  author = {Liu, Samuel and Kim, Youngsun and Matak, Martin and Lee, Gilwoo},
  title = {Getting Out and Getting Back: World and Behavior Grounding in {Real2Sim2Real} Co-Training},
  journal = {arXiv preprint arXiv:2610.00821},
  year = {2026}
}

@article{zhang2025softbodyeval,
  author = {Zhang, Kaifeng and Sha, Shuo and Jiang, Hanxiao and Loper, Matthew and
            Song, Hyunjong and Cai, Guangyan and Xu, Zhuo and Hu, Xiaochen and
            Zheng, Changxi and Li, Yunzhu},
  title = {Real-to-Sim Robot Policy Evaluation with Gaussian Splatting Simulation of Soft-Body Interactions},
  journal = {arXiv preprint arXiv:2511.04665},
  year = {2025}
}

@article{wang2025embodiedreamer,
  author = {Wang, Boyuan and Meng, Xinpan and Wang, Xiaofeng and Zhu, Zheng and
            Ye, Angen and Wang, Yang and Yang, Zhiqin and Ni, Chaojun and
            Huang, Guan and Wang, Xingang},
  title = {{EmbodieDreamer}: Advancing {Real2Sim2Real} Transfer for Policy Training via Embodied World Modeling},
  journal = {arXiv preprint arXiv:2507.05198},
  year = {2025}
}

@inproceedings{meng2026scenegen,
  author = {Meng, Yanxu and Wu, Haoning and Zhang, Ya and Xie, Weidi},
  title = {{SceneGen}: Single-Image {3D} Scene Generation in One Feedforward Pass},
  booktitle = {International Conference on 3D Vision},
  year = {2026}
}

@inproceedings{huang2025midi,
  author = {Huang, Zehuan and Guo, Yuan-Chen and An, Xingqiao and Yang, Yunhan and
            Li, Yangguang and Zou, Zi-Xin and Liang, Ding and Liu, Xihui and
            Cao, Yan-Pei and Sheng, Lu},
  title = {{MIDI}: Multi-Instance Diffusion for Single Image to {3D} Scene Generation},
  booktitle = {Proceedings of the IEEE/CVF Conference on Computer Vision and Pattern Recognition},
  pages = {23646--23657},
  year = {2025}
}

@inproceedings{zadaianchuk2026recgen,
  author = {Zadaianchuk, Andrii and Barcellona, Leonardo and Schuenemann, Lennard and
            Gumbsch, Christian and Wang, Zehao and Irshad, Muhammad Zubair and
            Despinoy, Fabien and Aljundi, Rahaf and Gavves, Stratis and Zakharov, Sergey},
  title = {Reconstruction by Generation: {3D} Multi-Object Scene Reconstruction from Sparse Observations},
  booktitle = {European Conference on Computer Vision},
  pages = {498--519},
  year = {2026}
}

@inproceedings{chopra2026physgs,
  author = {Chopra, Samarth and Liang, Jing and Seneviratne, Gershom and Manocha, Dinesh},
  title = {{PhysGS}: Bayesian-Inferred Gaussian Splatting for Physical Property Estimation},
  booktitle = {Proceedings of the IEEE/CVF Conference on Computer Vision and Pattern Recognition},
  pages = {18980--18990},
  year = {2026}
}

@inproceedings{cao2026physxanything,
  author = {Cao, Ziang and Hong, Fangzhou and Chen, Zhaoxi and Pan, Liang and Liu, Ziwei},
  title = {{PhysX-Anything}: Simulation-Ready Physical {3D} Assets from Single Image},
  booktitle = {Proceedings of the IEEE/CVF Conference on Computer Vision and Pattern Recognition},
  year = {2026}
}

@article{pfaff2025scalable,
  author = {Pfaff, Nicholas and Fu, Evelyn and Binagia, Jeremy and Isola, Phillip and Tedrake, Russ},
  title = {Scalable {Real2Sim}: Physics-Aware Asset Generation Via Robotic Pick-and-Place Setups},
  journal = {arXiv preprint arXiv:2503.00370},
  year = {2025}
}

@inproceedings{silberman2012support,
  author = {Silberman, Nathan and Hoiem, Derek and Kohli, Pushmeet and Fergus, Rob},
  title = {Indoor Segmentation and Support Inference from {RGBD} Images},
  booktitle = {European Conference on Computer Vision},
  pages = {746--760},
  year = {2012}
}

@article{yao2025cast,
  author = {Yao, Kaixin and Zhang, Longwen and Yan, Xinhao and Zeng, Yan and Zhang, Qixuan and
            Xu, Lan and Yang, Wei and Gu, Jiayuan and Yu, Jingyi},
  title = {{CAST}: Component-Aligned {3D} Scene Reconstruction from an {RGB} Image},
  journal = {ACM Transactions on Graphics},
  volume = {44},
  number = {4},
  year = {2025}
}

@article{zhang2026physomni,
  author = {Zhang, Xin and Chen, Yabo and Fang, Yijie and Qu, Wanying and Huang, Haibin and
            Zhang, Chi and Xu, Feng and Li, Xuelong},
  title = {{PhysOmni}: Physics-Grounded Multi-Object Scene Generation from a Single Image with Real-Time Interaction},
  journal = {arXiv preprint arXiv:2605.20290},
  year = {2026}
}

@inproceedings{huang2026sparcs,
  author = {Huang, Wei-Cheng and Han, Jiaheng and Ye, Xiaohan and Pan, Zherong and Hauser, Kris},
  title = {Simulation-Ready Cluttered Scene Estimation via Physics-Aware Joint Shape and Pose Optimization},
  booktitle = {Proceedings of Robotics: Science and Systems},
  year = {2026},
  doi = {10.15607/RSS.2026.XXII.174}
}

@inproceedings{yang2026robolab,
  author = {Yang, Jenai Xuning and Dagli, Rishit and Zook, Alex and Hadfield, Hugo and Goyal, Ankit and
            Birchfield, Stan and Ramos, Fabio and Tremblay, Jonathan},
  title = {{RoboLab}: A High-Fidelity Simulation Benchmark for Analysis of Task Generalist Policies},
  booktitle = {Proceedings of Robotics: Science and Systems},
  year = {2026}
}

@article{xiang2026functionpreserving,
  author = {Xiang, Tianyi and Xie, Xupeng and Cao, Jiahang and Luo, Andrew F. and Li, Haoang and Ma, Jun},
  title = {Function-Preserving Data Generation for Zero-Shot Real-to-Sim-to-Real Manipulation},
  journal = {arXiv preprint arXiv:2609.18293},
  year = {2026}
}

@article{tsui2026frontieragents,
  author = {Tsui, Brian Y.},
  title = {Measuring Frontier Code Agents at Digital-Twin Scene Construction},
  journal = {Preprints.org},
  year = {2026},
  note = {Version 2},
  doi = {10.20944/preprints202609.0344.v2},
  url = {https://www.preprints.org/manuscript/202609.0344/v2}
}

@article{chen2026agenticreal2sim,
  author = {Chen, Guanxiong and Xia, Qianjun and Peng, Jiawei and Zhang, Heng and Jing, Pengyu and
            Ma, Bole and Qian, Justin and Cheng, Yixian and Jiao, Ziyi and Zhou, Bingyang and
            Qu, Yiduo and Ye, Luoxin and Zhang, Kaifeng and Wang, Kunyi and Zeng, Weijia and
            Chen, Yunuo and Yang, Pengzhi and Zeng, Ziqiu and Luo, Siyuan and Wang, Huamin and
            Liu, Chao and Yuille, Alan and Shi, Fan and Zheng, Changxi and Li, Yunzhu and
            Jiang, Chenfanfu and Chen, Peter Yichen},
  title = {Agentic {Real2Sim}: Physics-Based World Modeling with Vision-Language Agents},
  journal = {arXiv preprint arXiv:2607.19190},
  year = {2026}
}

@inproceedings{mildenhall2020nerf,
  author = {Mildenhall, Ben and Srinivasan, Pratul P. and Tancik, Matthew and
            Barron, Jonathan T. and Ramamoorthi, Ravi and Ng, Ren},
  title = {{NeRF}: Representing Scenes as Neural Radiance Fields for View Synthesis},
  booktitle = {European Conference on Computer Vision},
  pages = {405--421},
  year = {2020}
}

@article{kerbl2023gaussians,
  author = {Kerbl, Bernhard and Kopanas, Georgios and Leimk{\"u}hler, Thomas and Drettakis, George},
  title = {{3D} Gaussian Splatting for Real-Time Radiance Field Rendering},
  journal = {ACM Transactions on Graphics},
  volume = {42},
  number = {4},
  year = {2023}
}

@article{chopra2024agrinerf,
  author = {Chopra, Samarth and Cladera, Fernando and Murali, Varun and Kumar, Vijay},
  title = {{AgriNeRF}: Neural Radiance Fields for Agriculture in Challenging Lighting Conditions},
  journal = {arXiv preprint arXiv:2409.15487},
  year = {2024}
}

@article{chopra2025splatblox,
  author = {Chopra, Samarth and Liang, Jing and Seneviratne, Gershom and Lee, Yonghan and
            Choi, Jaehoon and An, Jianyu and Cheng, Stephen and Manocha, Dinesh},
  title = {{Splatblox}: Traversability-Aware Gaussian Splatting for Outdoor Robot Navigation},
  journal = {arXiv preprint arXiv:2511.18525},
  year = {2025}
}

@inproceedings{szymanowicz2025bolt3d,
  author = {Szymanowicz, Stanislaw and Zhang, Jason Y. and Srinivasan, Pratul and Gao, Ruiqi and
            Brussee, Arthur and Holynski, Aleksander and Martin-Brualla, Ricardo and
            Barron, Jonathan T. and Henzler, Philipp},
  title = {{Bolt3D}: Generating {3D} Scenes in Seconds},
  booktitle = {Proceedings of the IEEE/CVF International Conference on Computer Vision},
  pages = {24846--24857},
  year = {2025}
}

@article{bahmani2025lyra,
  author = {Bahmani, Sherwin and Shen, Tianchang and Ren, Jiawei and Huang, Jiahui and
            Jiang, Yifeng and Turki, Haithem and Tagliasacchi, Andrea and Lindell, David B. and
            Gojcic, Zan and Fidler, Sanja and Ling, Huan and Gao, Jun and Ren, Xuanchi},
  title = {{Lyra}: Generative {3D} Scene Reconstruction via Video Diffusion Model Self-Distillation},
  journal = {arXiv preprint arXiv:2509.19296},
  year = {2025}
}

@inproceedings{yu2025metascenes,
  author = {Yu, Huangyue and Jia, Baoxiong and Chen, Yixin and Yang, Yandan and Li, Puhao and
            Su, Rongpeng and Li, Jiaxin and Li, Qing and Liang, Wei and Zhu, Song-Chun and
            Liu, Tengyu and Huang, Siyuan},
  title = {{MetaScenes}: Towards Automated Replica Creation for Real-World {3D} Scans},
  booktitle = {Proceedings of the IEEE/CVF Conference on Computer Vision and Pattern Recognition},
  year = {2025}
}

@article{hsu2026assetcentric,
  author = {Hsu, Christopher D. and Chaudhari, Pratik},
  title = {Asset-Centric Metric-Semantic Maps of Indoor Environments},
  journal = {arXiv preprint arXiv:2510.10778},
  year = {2026}
}

@inproceedings{xiang2025trellis,
  author = {Xiang, Jianfeng and Lv, Zelong and Xu, Sicheng and Deng, Yu and Wang, Ruicheng and
            Zhang, Bowen and Chen, Dong and Tong, Xin and Yang, Jiaolong},
  title = {Structured {3D} Latents for Scalable and Versatile {3D} Generation},
  booktitle = {Proceedings of the IEEE/CVF Conference on Computer Vision and Pattern Recognition},
  pages = {21469--21480},
  year = {2025}
}

@inproceedings{ansari2026messykitchens,
  author = {Ansari, Junaid Ahmed and Ding, Ran and Pizzati, Fabio and Laptev, Ivan},
  title = {{MessyKitchens}: Contact-Rich Object-Level {3D} Scene Reconstruction},
  booktitle = {European Conference on Computer Vision},
  pages = {520--538},
  doi = {10.1007/978-3-032-37574-2_29},
  year = {2026}
}

@inproceedings{yang2025instascene,
  author = {Yang, Zesong and Yang, Bangbang and Dong, Wenqi and Cao, Chenxuan and
            Cui, Liyuan and Ma, Yuewen and Cui, Zhaopeng and Bao, Hujun},
  title = {{InstaScene}: Towards Complete {3D} Instance Decomposition and Reconstruction from Cluttered Scenes},
  booktitle = {Proceedings of the IEEE/CVF International Conference on Computer Vision},
  pages = {7771--7781},
  year = {2025}
}

@inproceedings{ni2025dprecon,
  author = {Ni, Junfeng and Liu, Yu and Lu, Ruijie and Zhou, Zirui and Zhu, Song-Chun and
            Chen, Yixin and Huang, Siyuan},
  title = {Decompositional Neural Scene Reconstruction with Generative Diffusion Prior},
  booktitle = {Proceedings of the IEEE/CVF Conference on Computer Vision and Pattern Recognition},
  pages = {6022--6033},
  year = {2025}
}

@article{wu2026gator,
  author = {Wu, Qirui and Birchfield, Stan and Rabeti, Hesam and Chang, Angel X. and Wen, Bowen},
  title = {{GATOR}: Generative and Agentic {3D} Object Reconstruction from Casual Images},
  journal = {arXiv preprint arXiv:2610.11215},
  year = {2026}
}

@inproceedings{chen2024urdformer,
  author = {Chen, Zoey and Walsman, Aaron and Memmel, Marius and Mo, Kaichun and Fang, Alex and
            Vemuri, Karthikeya and Wu, Alan and Fox, Dieter and Gupta, Abhishek},
  title = {{URDFormer}: A Pipeline for Constructing Articulated Simulation Environments from Real-World Images},
  booktitle = {Proceedings of Robotics: Science and Systems},
  doi = {10.15607/RSS.2024.XX.124},
  year = {2024}
}

@inproceedings{xia2025drawer,
  author = {Xia, Hongchi and Su, Entong and Memmel, Marius and Jain, Arhan and Yu, Raymond and
            Mbiziwo-Tiapo, Numfor and Farhadi, Ali and Gupta, Abhishek and Wang, Shenlong and Ma, Wei-Chiu},
  title = {{DRAWER}: Digital Reconstruction and Articulation with Environment Realism},
  booktitle = {Proceedings of the IEEE/CVF Conference on Computer Vision and Pattern Recognition},
  pages = {21771--21782},
  year = {2025}
}

@article{wu2026urdfanything,
  author = {Wu, Zhuangzhe and Xin, Yue and Hou, Chengkai and Chen, Minghao and
            Lyu, Yaoxu and Zhang, Jieyu and Zhang, Shanghang},
  title = {{URDF-Anything+}: End-to-End Generation for Simulation-Ready Articulated Assets},
  journal = {arXiv preprint arXiv:2603.14010},
  year = {2026}
}

@inproceedings{xie2024physgaussian,
  author = {Xie, Tianyi and Zong, Zeshun and Qiu, Yuxing and Li, Xuan and Feng, Yutao and
            Yang, Yin and Jiang, Chenfanfu},
  title = {{PhysGaussian}: Physics-Integrated {3D} Gaussians for Generative Dynamics},
  booktitle = {Proceedings of the IEEE/CVF Conference on Computer Vision and Pattern Recognition},
  year = {2024}
}

@article{yang2025physworld,
  author = {Yang, Yu and Zhang, Zhilu and Zhang, Xiang and Zeng, Yihan and Li, Hui and Zuo, Wangmeng},
  title = {{PhysWorld}: From Real Videos to World Models of Deformable Objects via Physics-Aware Demonstration Synthesis},
  journal = {arXiv preprint arXiv:2510.21447},
  year = {2025}
}

@inproceedings{bianchini2025vysics,
  author = {Bianchini, Bibit and Zhu, Minghan and Sun, Mengti and Jiang, Bowen and
            Taylor, Camillo J. and Posa, Michael},
  title = {{Vysics}: Object Reconstruction Under Occlusion by Fusing Vision and Contact-Rich Physics},
  booktitle = {Proceedings of Robotics: Science and Systems},
  doi = {10.15607/RSS.2025.XXI.034},
  year = {2025}
}

@inproceedings{guo2013support,
  author = {Guo, Ruiqi and Hoiem, Derek},
  title = {Support Surface Prediction in Indoor Scenes},
  booktitle = {Proceedings of the IEEE International Conference on Computer Vision},
  pages = {2144--2151},
  year = {2013}
}

@inproceedings{zheng2013beyond,
  author = {Zheng, Bo and Zhao, Yibiao and Yu, Joey C. and Ikeuchi, Katsushi and Zhu, Song-Chun},
  title = {Beyond Point Clouds: Scene Understanding by Reasoning Geometry and Physics},
  booktitle = {Proceedings of the IEEE Conference on Computer Vision and Pattern Recognition},
  pages = {3127--3134},
  doi = {10.1109/CVPR.2013.402},
  year = {2013}
}

@article{wang2026vlk,
  author = {Wang, Yen-Jen and Li, Jiaman and Chen, Sirui and Truong, Takara E. and Xu, Pei and
            Abbeel, Pieter and Duan, Rocky and Sreenath, Koushil and Kanazawa, Angjoo and
            Sferrazza, Carmelo and Shi, Guanya and Liu, Karen},
  title = {{VLK}: Learning Humanoid Loco-Manipulation from Synthetic Interactions in Reconstructed Scenes},
  journal = {arXiv preprint arXiv:2606.30645},
  year = {2026}
}

@article{kim2026molmospaces,
  author = {Kim, Yejin and Pumacay, Wilbert and Rayyan, Omar and Argus, Max and Han, Winson and
            VanderBilt, Eli and Salvador, Jordi and Deshpande, Abhay and Hendrix, Rose and
            Jauhri, Snehal and Liu, Shuo and Shafiullah, Nur Muhammad Mahi and Guru, Maya and
            Eftekhar, Ainaz and Farley, Karen and Clay, Donovan and Duan, Jiafei and Guru, Arjun and
            Wolters, Piper and Herrasti, Alvaro and Lee, Ying-Chun and Chalvatzaki, Georgia and
            Cui, Yuchen and Farhadi, Ali and Fox, Dieter and Krishna, Ranjay},
  title = {{MolmoSpaces}: A Large-Scale Open Ecosystem for Robot Navigation and Manipulation},
  journal = {arXiv preprint arXiv:2602.11337},
  year = {2026}
}

@article{deshpande2026molmobot,
  author = {Deshpande, Abhay and Guru, Maya and Hendrix, Rose and Jauhri, Snehal and Eftekhar, Ainaz and
            Tripathi, Rohun and Argus, Max and Salvador, Jordi and Fang, Haoquan and
            Wallingford, Matthew and Pumacay, Wilbert and Kim, Yejin and Pfeifer, Quinn and
            Lee, Ying-Chun and Wolters, Piper and Rayyan, Omar and Zhang, Mingtong and Duan, Jiafei and
            Farley, Karen and Han, Winson and Vanderbilt, Eli and Fox, Dieter and Farhadi, Ali and
            Chalvatzaki, Georgia and Shah, Dhruv and Krishna, Ranjay},
  title = {{MolmoB0T}: Large-Scale Simulation Enables Zero-Shot Manipulation},
  journal = {arXiv preprint arXiv:2603.16861},
  year = {2026}
}

@article{yu2026score,
  author = {Yu, Raymond and Huey, William and Mukadam, Mustafa and Nagabandi, Anusha and Gupta, Abhishek},
  title = {Support-Constrained {RL} Enables Real-World Policy Improvement without Real-World Experience},
  journal = {arXiv preprint arXiv:2606.27475},
  year = {2026}
}

@article{jiang2025gsworld,
  author = {Jiang, Guangqi and Chang, Haoran and Qiu, Ri-Zhao and Liang, Yutong and Ji, Mazeyu and
            Zhu, Jiyue and Dong, Zhao and Zou, Xueyan and Wang, Xiaolong},
  title = {{GSWorld}: Closed-Loop Photo-Realistic Simulation Suite for Robotic Manipulation},
  journal = {arXiv preprint arXiv:2510.20813},
  year = {2025}
}

@inproceedings{jia2025discoverse,
  author = {Jia, Yufei and Wang, Guangyu and Dong, Yuhang and Wu, Junzhe and Zeng, Yupei and Lin, Haonan and
            Wang, Zifan and Ge, Haizhou and Gu, Weibin and Ding, Kairui and Yan, Zike and Cheng, Yunjie and
            Li, Yue and Wang, Ziming and Li, Chuxuan and Sui, Wei and Shi, Lu and Tian, Guanzhong and
            Huang, Ruqi and Zhou, Guyue},
  title = {{DISCOVERSE}: Efficient Robot Simulation in Complex High-Fidelity Environments},
  booktitle = {IEEE/RSJ International Conference on Intelligent Robots and Systems},
  year = {2025}
}

@inproceedings{pfaff2026scenesmith,
  author = {Pfaff, Nicholas and Cohn, Thomas and Zakharov, Sergey and Cory, Rick and Tedrake, Russ},
  title = {{SceneSmith}: Agentic Generation of Simulation-Ready Indoor Scenes},
  booktitle = {Proceedings of the 43rd International Conference on Machine Learning},
  series = {Proceedings of Machine Learning Research},
  volume = {306},
  pages = {98499--98567},
  year = {2026}
}

@inproceedings{openx2024embodiment,
  author = {{Open X-Embodiment Collaboration}},
  title = {{Open X-Embodiment}: Robotic Learning Datasets and {RT-X} Models},
  booktitle = {IEEE International Conference on Robotics and Automation},
  year = {2024},
  doi = {10.1109/ICRA57147.2024.10611477},
  url = {https://robotics-transformer-x.github.io/}
}

@inproceedings{khazatsky2024droid,
  author = {Khazatsky, Alexander and Pertsch, Karl and Nair, Suraj and others},
  title = {{DROID}: A Large-Scale In-The-Wild Robot Manipulation Dataset},
  booktitle = {Proceedings of Robotics: Science and Systems},
  year = {2024},
  doi = {10.15607/RSS.2024.XX.120},
  url = {https://www.roboticsproceedings.org/rss20/p120.html}
}

@inproceedings{walke2023bridgedata,
  author = {Walke, Homer Rich and Black, Kevin and Zhao, Tony Z. and Vuong, Quan and
            Zheng, Chongyi and Hansen-Estruch, Philippe and He, Andre Wang and Myers, Vivek and
            Kim, Moo Jin and Du, Max and Lee, Abraham and Fang, Kuan and Finn, Chelsea and Levine, Sergey},
  title = {{BridgeData V2}: A Dataset for Robot Learning at Scale},
  booktitle = {Proceedings of the 7th Conference on Robot Learning},
  series = {Proceedings of Machine Learning Research},
  volume = {229},
  pages = {1723--1736},
  year = {2023},
  url = {https://proceedings.mlr.press/v229/walke23a.html}
}

@article{escontrela2025gaussgym,
  author = {Escontrela, Alejandro and Kerr, Justin and Allshire, Arthur and Frey, Jonas and
            Duan, Rocky and Sferrazza, Carmelo and Abbeel, Pieter},
  title = {{GaussGym}: An Open-Source Real-to-Sim Framework for Learning Locomotion from Pixels},
  journal = {arXiv preprint arXiv:2510.15352},
  year = {2025},
  url = {https://arxiv.org/abs/2510.15352}
}

@inproceedings{black2025pi0,
  author = {Black, Kevin and Brown, Noah and Driess, Danny and Esmail, Adnan and
            Equi, Michael Robert and Finn, Chelsea and Fusai, Niccolo and Groom, Lachy and
            Hausman, Karol and Ichter, Brian and Jakubczak, Szymon and Jones, Tim and
            Ke, Liyiming and Levine, Sergey and Li-Bell, Adrian and Mothukuri, Mohith and
            Nair, Suraj and Pertsch, Karl and Shi, Lucy Xiaoyang and Smith, Laura and
            Tanner, James and Vuong, Quan and Walling, Anna and Wang, Haohuan and Zhilinsky, Ury},
  title = {{$\pi_0$}: A Vision-Language-Action Flow Model for General Robot Control},
  booktitle = {Proceedings of Robotics: Science and Systems},
  year = {2025},
  doi = {10.15607/RSS.2025.XXI.010},
  url = {https://www.roboticsproceedings.org/rss21/p010.html}
}

@article{genesis2026genesisworld1,
  author = {{Genesis AI Team}},
  title = {The Role of Simulation in Scalable Robotics, {Genesis World 1.0}, and the Path Forward},
  journal = {Genesis AI Blog},
  month = {May},
  year = {2026},
  url = {https://www.genesis.ai/blog/the-role-of-simulation-in-scalable-robotics-genesis-world-10-and-the-path-forward}
}

@inproceedings{todorov2012mujoco,
  author = {Todorov, Emanuel and Erez, Tom and Tassa, Yuval},
  title = {{MuJoCo}: A Physics Engine for Model-Based Control},
  booktitle = {IEEE/RSJ International Conference on Intelligent Robots and Systems},
  pages = {5026--5033},
  year = {2012},
  doi = {10.1109/IROS.2012.6386109}
}

@inproceedings{zakka2025mujocoplayground,
  author = {Zakka, Kevin and Tabanpour, Baruch and Liao, Qiayuan and Haiderbhai, Mustafa and
            Holt, Samuel and Luo, Jing Yuan and Allshire, Arthur and Frey, Erik and
            Sreenath, Koushil and Kahrs, Lueder Alexander and Sferrazza, Carmelo and
            Tassa, Yuval and Abbeel, Pieter},
  title = {Demonstrating {MuJoCo Playground}},
  booktitle = {Proceedings of Robotics: Science and Systems},
  year = {2025},
  doi = {10.15607/RSS.2025.XXI.020},
  url = {https://www.roboticsproceedings.org/rss21/p020.html}
}

@inproceedings{laskey2017dart,
  author = {Laskey, Michael and Lee, Jonathan and Fox, Roy and Dragan, Anca and Goldberg, Ken},
  title = {{DART}: Noise Injection for Robust Imitation Learning},
  booktitle = {Proceedings of the 1st Annual Conference on Robot Learning},
  series = {Proceedings of Machine Learning Research},
  volume = {78},
  pages = {143--156},
  year = {2017},
  url = {https://proceedings.mlr.press/v78/laskey17a.html}
}

@inproceedings{xiang2018posecnn,
  author = {Xiang, Yu and Schmidt, Tanner and Narayanan, Venkatraman and Fox, Dieter},
  title = {{PoseCNN}: A Convolutional Neural Network for {6D} Object Pose Estimation in Cluttered Scenes},
  booktitle = {Proceedings of Robotics: Science and Systems},
  year = {2018},
  doi = {10.15607/RSS.2018.XIV.019}
}

@inproceedings{roberts2021hypersim,
  author = {Roberts, Mike and Ramapuram, Jason and Ranjan, Anurag and Kumar, Atulit and
            Bautista, Miguel Angel and Paczan, Nathan and Webb, Russ and Susskind, Joshua M.},
  title = {{Hypersim}: A Photorealistic Synthetic Dataset for Holistic Indoor Scene Understanding},
  booktitle = {Proceedings of the IEEE/CVF International Conference on Computer Vision},
  pages = {10912--10922},
  year = {2021}
}

@article{chen1992object,
  title   = {Object modelling by registration of multiple range images},
  author  = {Chen, Yang and Medioni, G{\'e}rard},
  journal = {Image and Vision Computing},
  volume  = {10},
  number  = {3},
  pages   = {145--155},
  year    = {1992}
}

@article{fischler1981random,
  title   = {Random sample consensus: a paradigm for model fitting with applications to image analysis and automated cartography},
  author  = {Fischler, Martin A. and Bolles, Robert C.},
  journal = {Communications of the ACM},
  volume  = {24},
  number  = {6},
  pages   = {381--395},
  year    = {1981}
}

@article{schulman2017ppo,
  author = {Schulman, John and Wolski, Filip and Dhariwal, Prafulla and Radford, Alec and Klimov, Oleg},
  title = {Proximal Policy Optimization Algorithms},
  journal = {arXiv preprint arXiv:1707.06347},
  year = {2017},
  url = {https://arxiv.org/abs/1707.06347}
}

@article{raffin2021stablebaselines3,
  author = {Raffin, Antonin and Hill, Ashley and Gleave, Adam and Kanervisto, Anssi and Ernestus, Maximilian and Dormann, Noah},
  title = {{Stable-Baselines3}: Reliable Reinforcement Learning Implementations},
  journal = {Journal of Machine Learning Research},
  volume = {22},
  number = {268},
  pages = {1--8},
  year = {2021},
  url = {https://jmlr.org/papers/v22/20-1364.html}
}
}

\clearpage
\appendix
\maketitlesupplementary
\begingroup

\section{Additional Method Details}
\label{sec:supp-method-details}

\subsection{Scene inventory and support-graph inference [Sec.~\ref{sec:scene-evidence}]}
\label{sec:supp-scene-evidence}

The inventory request identifies objects to reconstruct from the source
RGB image. For each object, the VLM returns a short description, such as ``the large
ceramic cup,'' and a bounding box enclosing it in the image. We use the
description and box to prompt SAM~3 for the object's instance mask. Each
object also receives an ID. After segmentation, a separate request
infers direct support relationships from the original image, a copy
annotated with mask outlines and IDs, and the object labels.

\noindent\textbf{Prompts and inputs.}
Figure~\ref{fig:supp-prompts} summarizes the essential instructions for
the tabletop example. Both requests use \texttt{gpt-6.1-sol} with high
reasoning effort and structured JSON output. Inventory inference uses
separate system and user messages; graph inference uses a user message
with the source image, a copy annotated with object masks and IDs,
and the object labels.

\begingroup
\definecolor{promptBorder}{HTML}{9BBEB4}
\definecolor{promptBackground}{HTML}{F2F7F5}
\definecolor{graphPromptBorder}{HTML}{B8A2C8}
\definecolor{graphPromptBackground}{HTML}{F4EFF7}
\lstdefinestyle{sceneEvidence}{
  basicstyle=\ttfamily\small,
  columns=fullflexible,keepspaces=true,showstringspaces=false,
  breaklines=true,breakatwhitespace=true,
  frame=single,rulecolor=\color{promptBorder},
  backgroundcolor=\color{promptBackground},
  framesep=5pt,xleftmargin=5pt,xrightmargin=5pt,
  aboveskip=8pt,belowskip=8pt,captionpos=t
}
\begin{figure*}[t]
  \centering
  \begingroup
  \setlength{\fboxsep}{7pt}
  \setlength{\fboxrule}{0.5pt}
  \begin{minipage}[t]{.485\textwidth}
    \vspace{0pt}
    \fcolorbox{promptBorder}{promptBackground}{%
    \begin{minipage}[t]{\dimexpr\linewidth-2\fboxsep-2\fboxrule\relax}
      \small
      \textbf{(a) Scene inventory}\par\smallskip
      \textit{Input: source RGB image.}\par\smallskip
      \textbf{System.} You are a precise scene inventory system.
      Return only JSON conforming to
      the supplied schema.\par\smallskip
      \textbf{User.}
      \begin{itemize}[nosep,leftmargin=1.1em]
        \item Select at most 20 complete, independent primary objects
        relevant to robot interaction. Add their visible supports and
        the objects supporting them, up to 40 entities in total.
        These are limits, not quotas.
        \item Keep independently movable objects separate and rigidly
        attached parts together. Do not select both an object and its
        parts. Exclude people, robots, and unrelated background objects.
        \item Give each entity a unique \texttt{scene\_id}, label,
        a short identifying description, and tight bounding-box corners
        $(x_{\min},y_{\min},x_{\max},y_{\max})$, with integer coordinates
        scaled to 0--1000. Record visible parts.
        \item Mark primary/support roles and classify entities as
        \texttt{dynamic\_rigid}, \texttt{fixed\_structure}, or
        \texttt{world\_surface}.
      \end{itemize}
    \end{minipage}}
  \end{minipage}\hfill
  \begin{minipage}[t]{.485\textwidth}
    \vspace{0pt}
    \fcolorbox{graphPromptBorder}{graphPromptBackground}{%
    \begin{minipage}[t]{\dimexpr\linewidth-2\fboxsep-2\fboxrule\relax}
      \small
      \textbf{(b) Support graph}\par\smallskip
      \textit{Input: source image, a copy annotated with object masks and
      IDs, and object labels.}\par\smallskip
      \textbf{User.} Infer direct relationships using only the supplied
      object IDs. Outlines identify objects, not contact locations.
      \begin{itemize}[nosep,leftmargin=1.1em]
        \item \texttt{rests\_on}: carries some of the object's weight;
        \texttt{leans\_against}: provides a sideways brace;
        \texttt{A inside B}: A is contained within B;
        \texttt{touches}: contact without clear support or bracing.
        Containment alone does not imply support.
        \item For \texttt{A rests\_on B}, direct the edge from A to B.
        Report immediate supporters, not ancestors. Allow multiple
        supporters and keep \texttt{rests\_on} chains acyclic.
        \item Proximity or overlapping outlines alone do not establish
        contact. Give brief visual evidence and mark visibility as
        \texttt{visible}, \texttt{inferred}, or \texttt{unclear}.
        Use \texttt{none} when no relationship applies.
        \item Use \texttt{world} only for direct, visibly supported
        floor/ground relationships, never as a substitute for a tabletop.
        \item Return one JSON graph with \texttt{status} and
        \texttt{relationships} for each object. Each relationship contains
        \texttt{target\_id}, \texttt{mode}, \texttt{visibility}, and
        \texttt{evidence}.
      \end{itemize}
    \end{minipage}}
  \end{minipage}
  \endgroup
  \caption{\textbf{Inventory and support-graph VLM prompts.}
  Essential instructions for the MessyKitchens example, condensed for
  presentation. Inventory inference identifies reconstruction targets
  and provides segmentation prompts; a separate
  graph request uses the source image and its annotated copy to infer
  relationships between the identified objects.}
  \label{fig:supp-prompts}
\end{figure*}

\noindent\textbf{Example: a spoon spanning two dishes.}
Figure~\ref{fig:supp-scene-evidence} shows a MessyKitchens example from
the source image and instance masks to the inferred support graph and
final simulated scene. Its inventory contains nine entities, listed in
Table~\ref{tab:supp-scene-ids}. The inventory model assigns a descriptive
\texttt{scene\_id}. Preprocessing adds an \texttt{object\_id} and mask
filename. The spoon's scene ID is \texttt{wooden\_spoon} and its object
ID is \texttt{object\_7}.
The \hyperref[lst:supp-inventory-output]{\emph{inventory output}} shows
the spoon's identity, description, and image bounding box. The
\hyperref[lst:supp-graph-output]{\emph{support-graph output}} shows its
complete graph entry, with both the round plate and floral dish as direct
supporters. Seating and pose search use these graph relationships.
These relationships are inferred from the image; the
\texttt{visible} labels are model judgments, not ground-truth annotations.

\begin{figure*}[htbp]
  \centering
  \includegraphics[width=\linewidth]{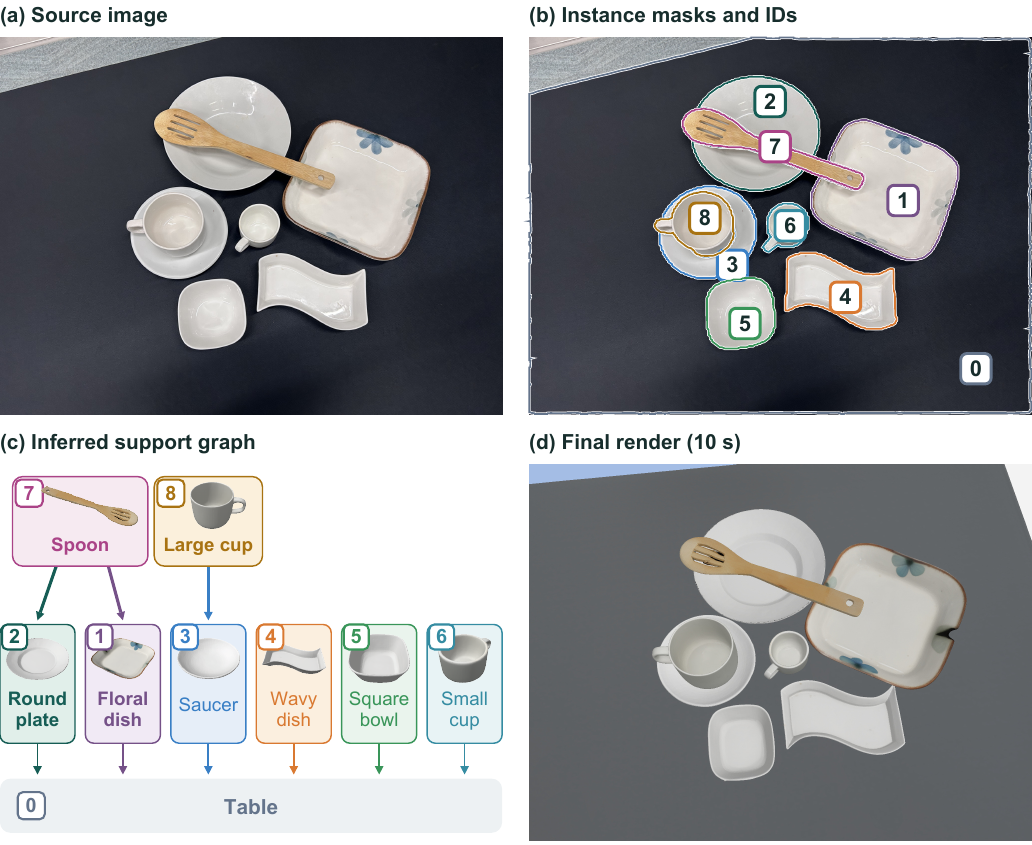}
  \caption{\textbf{Single-view reconstruction example from MessyKitchens.}
  (a) Source image. (b) SAM~3 masks and object IDs. (c) Inferred support
  graph, with arrows from objects to their supports; spoon 7 rests on
  plate 2 and floral dish 1. (d) Source-camera render after 10 seconds of
  simulation. IDs 0--8 denote \texttt{object\_0}--\texttt{object\_8}; mask
  outlines identify objects, not contact locations.}
  \label{fig:supp-scene-evidence}
\end{figure*}

\begin{table*}[htbp]
  \centering
  \small
  \caption{\textbf{Saved inventory for the MessyKitchens scene in Fig.~\ref{fig:supp-scene-evidence}.} Object IDs are assigned during
  preprocessing; scene IDs and labels are returned by the inventory model.}
  \label{tab:supp-scene-ids}
  \begin{tabular}{lll}
    \toprule
    Object ID & Scene ID & Label \\
    \midrule
    \texttt{object\_0} & \texttt{table} & black table \\
    \texttt{object\_1} & \texttt{floral\_dish} & square floral ceramic dish \\
    \texttt{object\_2} & \texttt{round\_plate} & round ceramic plate \\
    \texttt{object\_3} & \texttt{saucer} & white saucer \\
    \texttt{object\_4} & \texttt{wavy\_dish} & wavy rectangular ceramic dish \\
    \texttt{object\_5} & \texttt{small\_square\_bowl} & small square ceramic bowl \\
    \texttt{object\_6} & \texttt{small\_cup} & small ceramic cup \\
    \texttt{object\_7} & \texttt{wooden\_spoon} & wooden slotted spoon \\
    \texttt{object\_8} & \texttt{large\_cup} & large ceramic cup \\
    \bottomrule
  \end{tabular}
\end{table*}

\captionsetup[lstlisting]{labelformat=empty,labelsep=none}
\begin{lstlisting}[style=sceneEvidence,
  caption={Inventory output for spoon 7 in
  Fig.~\ref{fig:supp-scene-evidence}.},
  label={lst:supp-inventory-output}]
{
  "scene_id": "wooden_spoon",
  "label": "wooden slotted spoon",
  "referring_expression": "the long wooden slotted spoon lying diagonally across the round plate toward the floral dish",
  "bbox_xyxy_1000": [
    305,
    189,
    667,
    405
  ]
}
\end{lstlisting}
\begin{lstlisting}[style=sceneEvidence,
  caption={Support-graph output for spoon 7 in
  Fig.~\ref{fig:supp-scene-evidence}: round plate 2 and floral dish 1
  are its supports.},
  label={lst:supp-graph-output}]
{
  "object_7": {
    "status": "resolved",
    "relationships": [
      {
        "target_id": "object_2",
        "mode": "rests_on",
        "visibility": "visible",
        "evidence": "The spoon's broad head lies across the round plate and its rim."
      },
      {
        "target_id": "object_1",
        "mode": "rests_on",
        "visibility": "visible",
        "evidence": "The handle bridges onto the floral dish, resting across its left rim."
      }
    ]
  }
}
\end{lstlisting}
\endgroup

\noindent\textbf{Graph consistency.}
A \texttt{rests\_on} edge points from the supported object to its supporter.
The prompt allows multiple direct supporters but requires these edges to
form an acyclic graph. After inference, we repeatedly remove one edge
from each detected cycle until no cycle remains. In this image-only step,
we prefer removing edges labeled \texttt{unclear}, then \texttt{inferred},
then \texttt{visible}, using object-ID order to break ties. The example contains
no cycle, so no edges are removed.

\subsection{Analytical roll selection for contact-pair proposals [Sec.~\ref{sec:contact-initialization}]}
\label{sec:supp-contact-roll}
Let $R_0$ align the object's sampled contact-pair direction with
$\hat{\mathbf d}_b$, the unit direction between the two sampled support
points. The remaining rotation $\phi$ is about this contact axis.
For contact $k$, let $\mathbf u_k$ and $\mathbf v_k$ be the projections of
the object's outward unit normal after rotation by $R_0$ and the negative
outward unit normal of its support onto the plane perpendicular to
$\hat{\mathbf d}_b$, respectively. We compute the roll angle analytically as
\begin{equation}
  \phi = \operatorname{atan2}\!\left(
    \sum_{k=1}^{2}\hat{\mathbf d}_b^\top(\mathbf u_k\times\mathbf v_k),
    \sum_{k=1}^{2}\mathbf u_k^\top\mathbf v_k
  \right).
  \label{eq:contact-normal-roll}
\end{equation}
This angle maximizes opposing normal alignment across both contacts.

\subsection{Preliminary ranking of initializer candidates [Sec.~\ref{sec:contact-initialization}]}
\label{sec:supp-initializer-ranking}
The initializer ranks candidate poses before detailed collision checks and
rendering to allocate its wall-clock budget $B_{\mathrm{init}}$.
For a candidate $T_\phi=(R_\phi,\mathbf t_\phi)$, we approximate the
silhouette using the convex hull of projected mesh points. Let
$\widetilde I_i(T_\phi)$ be its IoU with the source mask $M_i$. We rank by
\begin{equation}
  \begin{aligned}
    q_i(T_\phi) &= \lambda_I\bigl[1-\widetilde I_i(T_\phi)\bigr]
    +\lambda_n e_n(T_\phi)\\
    &\quad+\lambda_x\left(\frac{\delta_x(T_\phi)}{0.5L_i}\right)^2
    +\lambda_R\left(\frac{\delta_R(T_\phi)}{60^\circ}\right)^2,
  \end{aligned}
  \label{eq:initializer-priority}
\end{equation}
where lower scores are preferred and $L_i$ is the reconstructed mesh's
extent along its dominant principal axis. We use $\lambda_I=1$,
$\lambda_n=0.05$, and $\lambda_x=\lambda_R=0.1$, assigning the largest
weight to IoU to prioritize source fidelity. Here, $\delta_x$ and
$\delta_R$ measure mesh-centroid displacement and angular change relative
to the input pose. The normal mismatch is
$e_n(T_\phi)=\frac14\sum_{k=1}^{2}[1+(R_\phi\mathbf n_{a,k})^\top\mathbf n_{b,k}]$,
where $\mathbf n_{a,k}$ and $\mathbf n_{b,k}$ are outward unit normals in
the object and world frames, respectively. It is zero when the paired
surfaces have opposing normals. This preliminary score determines the
evaluation order; the final selection maximizes rendered source-mask IoU
among candidates passing the geometric and support checks.

\subsection{Refiner partner selection and search order [Sec.~\ref{sec:coarse-to-fine-refinement}]}
\label{sec:supp-refiner-partners}
Each support group contains a target $i$ and its direct supports; a
\emph{visit} is a short allocation of search time to that group.
The refiner first proposes updates to $i$ alone. If none is accepted,
it keeps the search scale and selects a related movable object $j$:
first, an object that should remain separate from $i$ but is touching or
too close to it; next, the support that carries $i$'s weight least
consistently; otherwise, an object resting on $i$.
Support consistency is measured by $\ell_{ij}$, the fraction of recorded
time points in the rollout's final second when $j$ carries load from $i$.

If attempts keep failing, the refiner tries moving $i$ and $j$ together,
then $j$ alone, then $i$, $j$, and $k$ together if $k$ is available.
Here, $k$ is another movable support of $i$ or an object resting on it.
Only the selected objects' initial poses change. If every attempt fails,
the refiner pauses the group until $i$ or a related object changes pose. An accepted update halves the translation
and rotation ranges down to the finest scale. In the spoon-and-rolling-pin
example, joint updates move the rolling pin together with the inverted
mug that supports it.

\noindent\textbf{Single-support fallback.}
If no target has multiple supports, we consider supported objects that
violate contact, motion, or penetration limits. We rank them by excluded
contact, support deficit, and the largest motion or penetration relative
to its limit, with IoU as the final tie-breaker.

\subsection{Parameter settings [Sec.~\ref{sec:implementation-details}]}
\label{sec:supp-parameters}
Table~\ref{tab:supp-parameters} consolidates the default simulation and
pose-search settings.
Search and validation limits are acceptance thresholds, not measured errors.

\section{Additional Experimental Results}
\label{sec:supp-additional-results}

\subsection{Runtime breakdown [Sec.~\ref{sec:reconstruction-benchmark}]}
Table~\ref{tab:supp-stage-timing} reports the time spent in each stage,
using the same timing boundaries as the reconstruction benchmark. Shared RGB
images, DA3-inferred depth, and object meshes are prepared before timing
begins. Construction and seating account for the largest share in all
four datasets because this stage builds and checks candidate layouts.
The additional runtime on HyperSim is concentrated mainly in this stage,
rather than in physical-property estimation or export.

\subsection{Refiner budget and compute allocation [Sec.~\ref{sec:ablation}]}
For each of the 16 ablation scenes, all budgets start from the same
settled layout produced by the initializer. We remove the global limits
on visits and candidate rollouts. When multi-support targets exist, we refine those
objects; single-support objects that violate contact, motion, or
penetration limits are considered only when no multi-support target exists.
Meshes, physical properties, cameras, loss weights, and acceptance
thresholds remain fixed. The budget covers search; the final
10-second validation is measured separately.

At 320 seconds, mean IoU is $0.50$ percentage points higher than at
60 seconds, with the same support count and nearly unchanged drift,
for roughly four times the measured search time
(Table~\ref{tab:supp-refiner-budget}). Figure~\ref{fig:supp-refiner-timeline}
shows the added compute going mainly to further rolling-pin refinement.
Search can finish early if no object meets these criteria or all groups
are paused after unsuccessful target-only and related-object moves.
A paused group is reconsidered only after its target or a related
object changes pose.

\subsection{Refiner loss terms [Sec.~\ref{sec:ablation}]}
We retain one loss term at its original weight and set the others to zero,
using the same starting layouts and a 60-second search budget
(Fig.~\ref{fig:supp-refiner-loss}). The acceptance checks remain active,
so a single-term objective cannot bypass limits on support loss, motion,
penetration, or source alignment. The radar plots show changes from the
starting layouts, allowing improvements in the selected metric to be
compared with their effects on the other metrics. The full objective is
shown in every panel as a common reference.

\subsection{Qualitative comparisons [Sec.~\ref{sec:qualitative-comparison}]}
Figures~\ref{fig:supp-qualitative-extended} and
\ref{fig:supp-hypersim-fullpage} show how the reconstructed arrangements
change after release. Support2Sim preserves the cup stack and the marker
on its brick, while the HyperSim crops highlight retained background
chairs and teacups. The gains are not uniform: SimFoundry has higher
final IoU in the example with mugs and a slotted spatula, whereas
Support2Sim realizes more supports. The dining room remains a failure case, with displaced
small objects and greater drift than SimFoundry.

\subsection{PyBullet and MuJoCo comparison [Sec.~\ref{sec:implementation-details}]}
\label{sec:supp-simulator-comparison}
We run three reconstructions for 10 seconds in both simulators
(Fig.~\ref{fig:supp-simulator-exports}). Initial poses, meshes, masses,
centers of mass, inertias, friction, and cameras match. Both use 240\,Hz
stepping, gravity of $9.8\,\mathrm{m/s^2}$, zero initial velocities, and
no body damping for this comparison.

We use each simulator's built-in contact solver. PyBullet iteratively
applies impulses to correct velocities and resist penetration and
sliding. MuJoCo's soft-contact model allows collision surfaces to overlap
slightly to represent compression while the objects remain rigid. The
solvers can produce different forces for the same poses and friction
coefficients. We settled the release poses in PyBullet. The additional
settling and slow sliding in MuJoCo are consistent with its different
contact model. Both simulators realize all intended supports in these
three examples.

\subsection{Support graph quality across VLMs [Sec.~\ref{sec:scene-evidence}]}
\label{sec:supp-support-graph-models}
Because the support graph guides object placement, we test how its
accuracy varies across VLMs and how consistently each model predicts the
same supports across trials. We evaluate GPT-6.1 Sol, GPT-6.2 Sol,
GPT-6.2 Astra, GPT-6.1 Terra, and GPT-6 Luna on the 16 ablation scenes,
with five trials per model and scene. Each trial uses the same source
image, object IDs, and prompt, with high reasoning effort. We curated a
ground-truth graph for each scene through human review. We score the
repaired graphs against this reference (Table~\ref{tab:supp-graph-model-comparison}).

\noindent\textbf{Cyclic supports.}
In the curated Jenga graph, following the support arrows leads back to
the starting block after eight steps. We use a directed acyclic graph
(DAG) to seat each support before the objects it carries. Keeping all
eight edges would require the starting block to be seated before itself.
All 25 predicted graphs omit at least one edge from the loop
(Fig.~\ref{fig:supp-graph-disagreements}b).

\twocolumn[%
\begin{minipage}{\textwidth}
  \centering
  \small
  \setlength{\tabcolsep}{4pt}
  \renewcommand{\arraystretch}{0.95}
  \captionof{table}{\textbf{Single-view reconstruction and pose-search parameters.}
  $L_i$ denotes the object's principal-axis length, $r_j$ the support
  footprint inradius, and pp percentage points. Candidate limits are
  scene-wide ceilings shared across support groups.}
  \label{tab:supp-parameters}
  \begin{tabular*}{\textwidth}{@{\extracolsep{\fill}}p{.34\textwidth}p{.32\textwidth}p{.29\textwidth}@{}}
    \toprule
    Parameter & Value & Scope / role \\
    \midrule
    \multicolumn{3}{@{}l}{\textcolor[HTML]{165D55}{\textbf{Scene evidence and simulation}}} \\
    Inventory and graph model & GPT-6.1 Sol; high reasoning & Both inference stages \\
    Inventory limits & 20 primary; 40 total & Tabletop prompt profile \\
    Simulator & PyBullet 3.2.7 & Rigid-body rollouts \\
    Physics integration rate; gravity & 240\,Hz; $9.8\,\mathrm{m/s^2}$ & PyBullet rollouts \\
    Solver iterations / internal substeps & 50 / 0 & Fixed-step simulation; real-time off \\
    Initial linear / angular velocity & 0 / 0 & Each rollout starts from rest \\
    Loaded collision margin & 1\,mm & Convex collision parts \\
    Lateral friction & $1.0$ & Movable rigid objects \\
    Restitution; rolling/spinning friction & $0$ & Movable rigid objects \\
    \midrule
    \multicolumn{3}{@{}l}{\textcolor[HTML]{165D55}{\textbf{Collision decomposition}}} \\
    Decomposition & V-HACD & Compound convex collision geometry \\
    Resolution / concavity & 200,000 / 0.01 & V-HACD settings \\
    Plane / hull downsampling & 4 / 4 & V-HACD settings \\
    Initial hull vertex limit & 32 & Before refitting \\
    Minimum hull-volume setting & 0.001 & V-HACD \texttt{minVolumePerCH} \\
    Refit collision parts to source mesh & Enabled & Rebuild hulls from assigned source triangles \\
    Collision-shape shrinkage & 0\,mm & Separate from the collision margin \\
    \midrule
    \multicolumn{3}{@{}l}{\textcolor[HTML]{165D55}{\textbf{Graph-guided seating}}} \\
    Support boundary margin & $\min(10\,\mathrm{mm},0.8r_j)$ & Projected center of mass \\
    Retained patch overlap & $\geq 90\%$ of the best overlap & Smallest admissible shift \\
    Vertical clearance & $0.1$\,mm & Before gravity settling \\
    \midrule
    \multicolumn{3}{@{}l}{\textcolor[HTML]{165D55}{\textbf{Contact-guided pose initializer}}} \\
    Wall-clock cap $B_{\mathrm{init}}$ & 60\,s & Per scene \\
    Surface-point limits & 96 object; 64 per support & Farthest-point selection \\
    Minimum pair span & $0.15L_i$ & Object contact pair \\
    Pair-span mismatch tolerance & $\min(0.01L_i,1\,\mathrm{mm})$ & Object/support pair matching \\
    Pose / geometry-check / render caps & 6,000 / 600 / 64 & Successive screening stages \\
    Support gap; penetration tolerance & 1\,mm; 0.1\,mm & Geometric feasibility \\
    Ranking weights $(\lambda_I,\lambda_n,\lambda_x,\lambda_R)$ & $(1,0.05,0.1,0.1)$ & Approximate ranking \\
    Pose-change normalizers & $0.5L_i$; $60^\circ$ & Translation; rotation \\
    \midrule
    \multicolumn{3}{@{}l}{\textcolor[HTML]{165D55}{\textbf{Coarse-to-fine refiner}}} \\
    Wall-clock cap $B_{\mathrm{ref}}$ & 60\,s & Per scene \\
    Global visit / rollout-count limits & 12 / 64 & Disabled in budget sweep \\
    Translation radii & $15\%,\ 7.5\%,\ 3.75\%$ of $L_i$ & Coarse to fine \\
    Rotation radii & $5^\circ,\ 2.5^\circ,\ 1.25^\circ$ & Coarse to fine \\
    Candidate / validation rollouts & 2\,s / 10\,s & Simulation duration \\
    Scoring interval; recording rate & Final 1\,s; 60\,Hz & Candidate-rollout scoring \\
    Loss weights $(w_I,w_S,w_M,w_P)$ & $(1,2,0.05,0.02)$ & IoU, supports, motion, penetration \\
    Motion / penetration normalizers & 2\,mm and $3^\circ$ / 0.1\,mm & Dimensionless losses \\
    Group / surrounding-object IoU tolerance & 0.5\,pp / 2\,pp & Source-fidelity guards \\
    Loss-improvement tolerance & $10^{-4}$ & Support-preserving acceptance \\
    Search penetration / motion limits & 1\,mm / 50\,mm and $30^\circ$ & Candidate acceptance \\
    Final displacement / rotation limits & 10\,mm / $10^\circ$ & Longer-rollout validation \\
    \bottomrule
  \end{tabular*}
\end{minipage}%
]

\begin{table*}[ht]
  \centering
  \small
  \caption{\textbf{Support2Sim runtime by stage.}
  Mean wall-clock minutes $\pm$ sample standard deviation after shared
  preparation of RGB images, DA3-inferred depth, and object meshes.}
  \label{tab:supp-stage-timing}
  \begin{tabular*}{\linewidth}{@{\extracolsep{\fill}}lcccc@{}}
    \toprule
    Stage & MessyKitchens & YCB-Video & Picasso & HyperSim \\
    \midrule
    FoundationPose                   & $0.38\pm0.04$ & $0.27\pm0.07$ & $0.26\pm0.07$ & $0.87\pm0.47$ \\
    Physical property estimation     & $1.87\pm0.13$ & $1.64\pm0.02$ & $1.67\pm0.07$ & $1.88\pm0.25$ \\
    Construction and seating         & $5.50\pm1.88$ & $2.32\pm0.55$ & $2.56\pm0.26$ & $13.17\pm6.54$ \\
    Downstream review and export     & $0.56\pm0.10$ & $0.22\pm0.03$ & $0.30\pm0.02$ & $0.75\pm0.15$ \\
    Other construction overhead     & $1.50\pm0.08$ & $1.47\pm0.00$ & $1.47\pm0.00$ & $1.52\pm0.03$ \\
    Fine search and validation       & $0.30\pm0.24$ & $0.13\pm0.04$ & $0.22\pm0.15$ & $0.63\pm0.35$ \\
    \midrule
    \textbf{Total} & $\mathbf{10.11\pm1.86}$ & $\mathbf{6.04\pm0.67}$ &
    $\mathbf{6.48\pm0.57}$ & $\mathbf{18.82\pm7.50}$ \\
    \bottomrule
  \end{tabular*}
\end{table*}

\begin{table*}[ht]
  \centering
  \small
  \caption{\textbf{Effect of longer refiner budgets on the 16 ablation scenes.}
  Each scene starts from the same settled initializer output at every
  budget. Search time is measured wall-clock time. We report source-mask
  IoU, realized supports, and drift from 10-second rollouts.
  \textcolor[HTML]{165D55}{\textbf{Bold teal}} marks the best performance
  values, including ties.}
  \label{tab:supp-refiner-budget}
  \newcommand{\budgetBest}[1]{\textcolor[HTML]{165D55}{\textbf{#1}}}
  \begin{tabular*}{\linewidth}{@{\extracolsep{\fill}}rrrrr@{}}
    \toprule
    Budget (s) & Mean search (s) & IoU (\%) $\uparrow$ &
    Supports $\uparrow$ & Drift (mm) $\downarrow$ \\
    \midrule
    0 & 0.0 & 52.57 & 121/129 & 0.191 \\
    15 & 12.7 & 52.41 & 122/129 & 0.192 \\
    30 & 24.7 & 53.25 & 122/129 & 0.171 \\
    60 & 47.2 & 53.69 & \budgetBest{123/129} & 0.172 \\
    90 & 67.9 & 53.84 & \budgetBest{123/129} & 0.173 \\
    120 & 86.3 & 54.06 & 122/129 & 0.180 \\
    180 & 122.6 & 53.99 & \budgetBest{123/129} & \budgetBest{0.167} \\
    240 & 152.7 & 54.05 & \budgetBest{123/129} & 0.174 \\
    320 & 193.0 & \budgetBest{54.18} & \budgetBest{123/129} & 0.171 \\
    \bottomrule
  \end{tabular*}
\end{table*}

\begin{figure*}[ht]
  \centering
  \includegraphics[width=.50\linewidth]{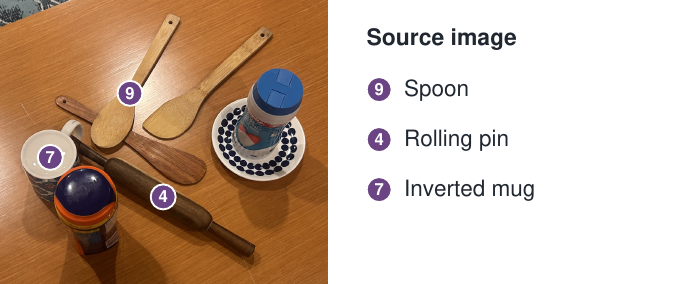}
  \par\smallskip
  \includegraphics[width=\linewidth]{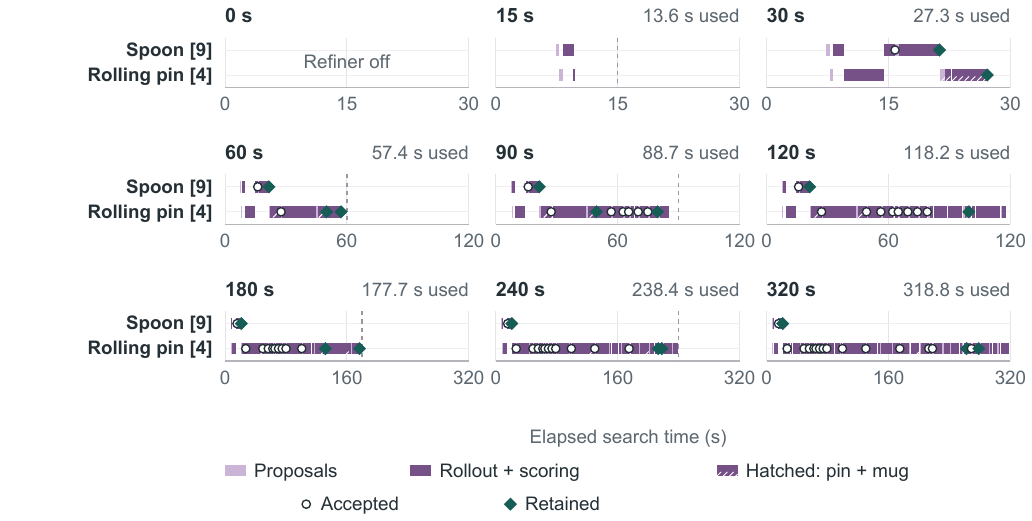}
  \caption{\textbf{Refiner compute on the spoon-and-rolling-pin scene.}
  The source labels the spoon (9), rolling pin (4), and mug (7).
  Panel titles give the allowed budgets; axes show elapsed search time.
  Light bars show time spent generating and geometrically checking poses;
  dark bars show time spent simulating and scoring them. Hatched intervals
  move the rolling pin and mug jointly within the rolling-pin group.
  Circles mark updates accepted
  during search; diamonds mark updates kept after final 10-second validation.}
  \label{fig:supp-refiner-timeline}
\end{figure*}

\begin{figure*}[t]
  \centering
  \includegraphics[width=\linewidth]{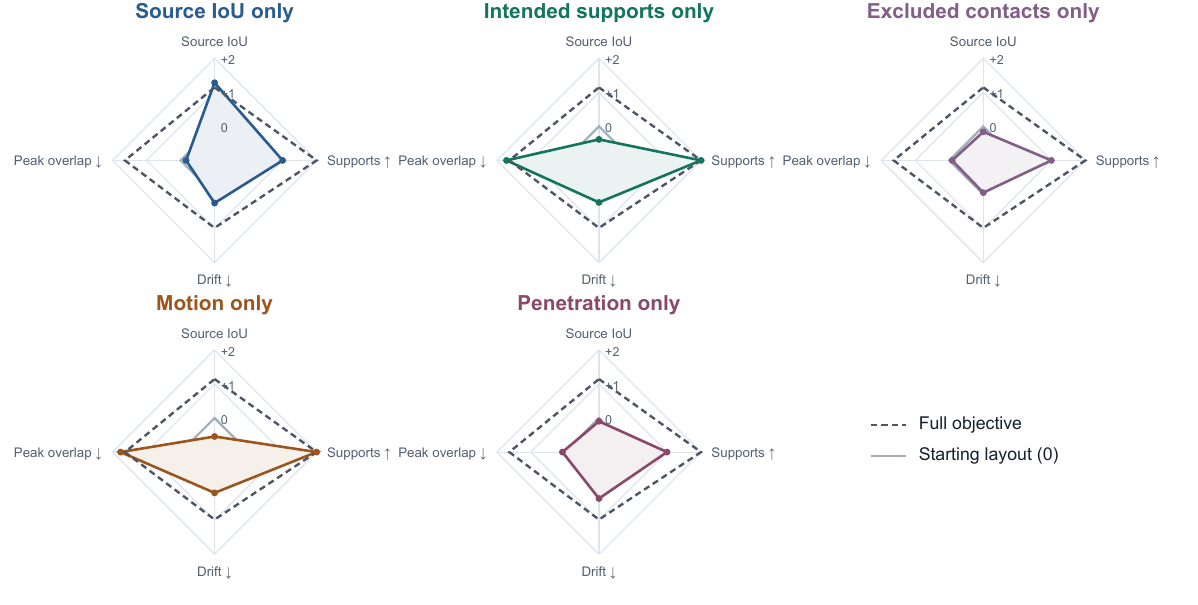}
  \caption{\textbf{Ablation of the refiner loss.}
  The full loss is $\mathcal{L}=w_I(1-\bar I)+w_S\mathcal{L}_S+
  w_M\mathcal{L}_M+w_P\mathcal{L}_P$
  (Sec.~\ref{sec:coarse-to-fine-refinement}), where $\bar I$ is mean
  source-mask IoU and $\mathcal{L}_S$, $\mathcal{L}_M$, and $\mathcal{L}_P$
  penalize unloaded intended supports, motion, and penetration.
  Each panel retains one term at its original weight and sets the others
  to zero; the dashed curve uses the full loss. All runs use the same
  16 scenes, a 60-second search cap, unchanged acceptance checks, and
  10-second evaluation rollouts. The zero ring represents the unrefined
  layouts. Farther outward means higher IoU and support counts, or lower
  drift and penetration. Each ring interval is one IoU percentage point,
  one realized support, $0.02$\,mm of mean drift, or $0.01$\,mm of mean
  peak penetration (the largest inter-object penetration during each
  rollout, averaged across scenes). Realized supports rise from 121/129
  to 123/129 for the full, supports-only, and motion-only losses, and to
  122/129 for the source-only and penetration-only losses.}
  \label{fig:supp-refiner-loss}
\end{figure*}

\begin{figure*}[p]
  \centering
  \begin{minipage}{\linewidth}
    \centering
    \small
    \captionof{table}{\textbf{Support graph quality and repeatability.}
    Scores are percentages, averaged over trials within each scene and then
    over scenes. F1 intervals are 95\% confidence intervals from resampling
    scenes. Exact multi-support accuracy requires the full support set to
    match and uses the 13 scenes with verified multi-support objects.
    For trials with directed support edge sets $E_a$ and $E_b$, Jaccard
    overlap is $J(E_a,E_b)=\frac{|E_a\cap E_b|}{|E_a\cup E_b|}$.
    Trial agreement averages $J$ over all 10 trial pairs per scene and
    then over scenes. Bold teal marks the highest mean for each metric.}
    \label{tab:supp-graph-model-comparison}
    \begin{tabular*}{\linewidth}{@{\extracolsep{\fill}}lccccc@{}}
      \toprule
      Model & Precision & Recall & F1 [95\% CI] & Exact multi-support & Trial agreement \\
      \midrule
      GPT-6.1 Sol & 97.3 & 96.0 & 96.4 [93.8, 98.6] & 77.1 & 95.9 \\
      GPT-6.2 Sol & 99.0 & 97.8 & 98.3 [97.0, 99.4] & \textcolor[HTML]{165D55}{\textbf{82.1}} & 98.0 \\
      GPT-6.2 Astra & \textcolor[HTML]{165D55}{\textbf{99.4}} & \textcolor[HTML]{165D55}{\textbf{98.6}} & \textcolor[HTML]{165D55}{\textbf{98.9}} [97.9, 99.8] & \textcolor[HTML]{165D55}{\textbf{82.1}} & \textcolor[HTML]{165D55}{\textbf{99.4}} \\
      GPT-6.1 Terra & 98.4 & 95.6 & 96.4 [93.4, 98.8] & 78.7 & 95.7 \\
      GPT-6 Luna & 93.7 & 84.0 & 86.6 [77.1, 94.3] & 49.2 & 93.7 \\
      \bottomrule
    \end{tabular*}
  \end{minipage}
  \par\medskip
  \includegraphics[width=\linewidth]{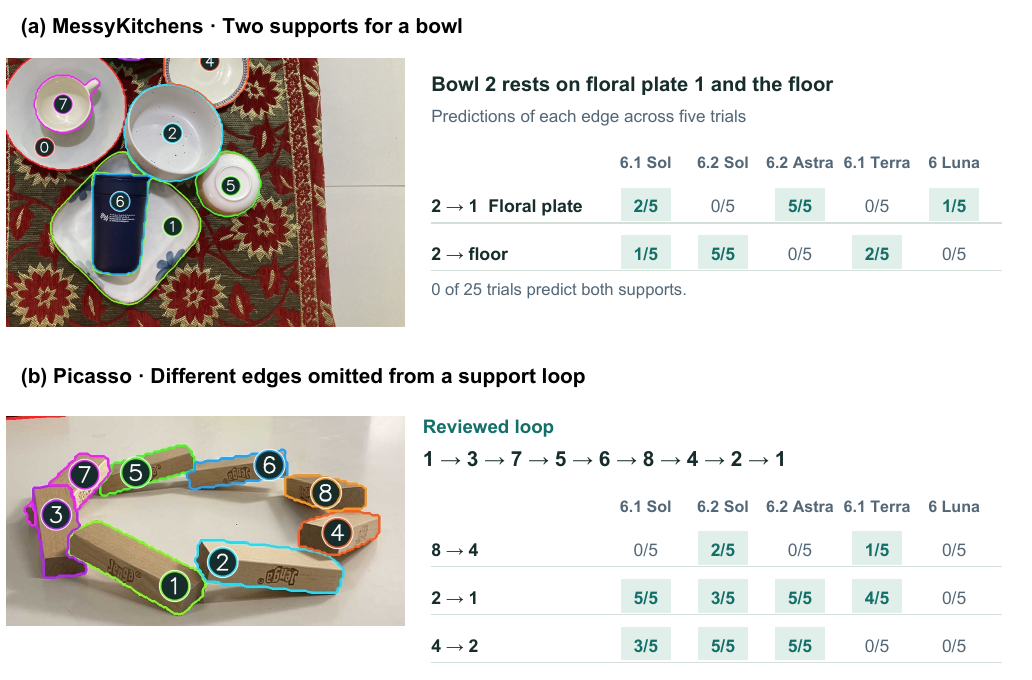}
  \caption{\textbf{Two support graph disagreements.}
  Arrows point from objects to their supports. Counts show how many of
  five trials predict each edge. (a) Bowl 2 rests partly on floral
  plate 1 and partly on the floor. Astra predicts the plate in all five
  trials, and GPT-6.2 Sol predicts the floor in all five. No trial
  predicts both. (b) Astra omits $8\to4$ in the Jenga loop. GPT-6.2 Sol
  includes it twice and omits $2\to1$ in those trials.}
  \label{fig:supp-graph-disagreements}
\end{figure*}

\clearpage

\begin{figure*}[p]
  \centering
  \includegraphics[page=1,width=\textwidth]{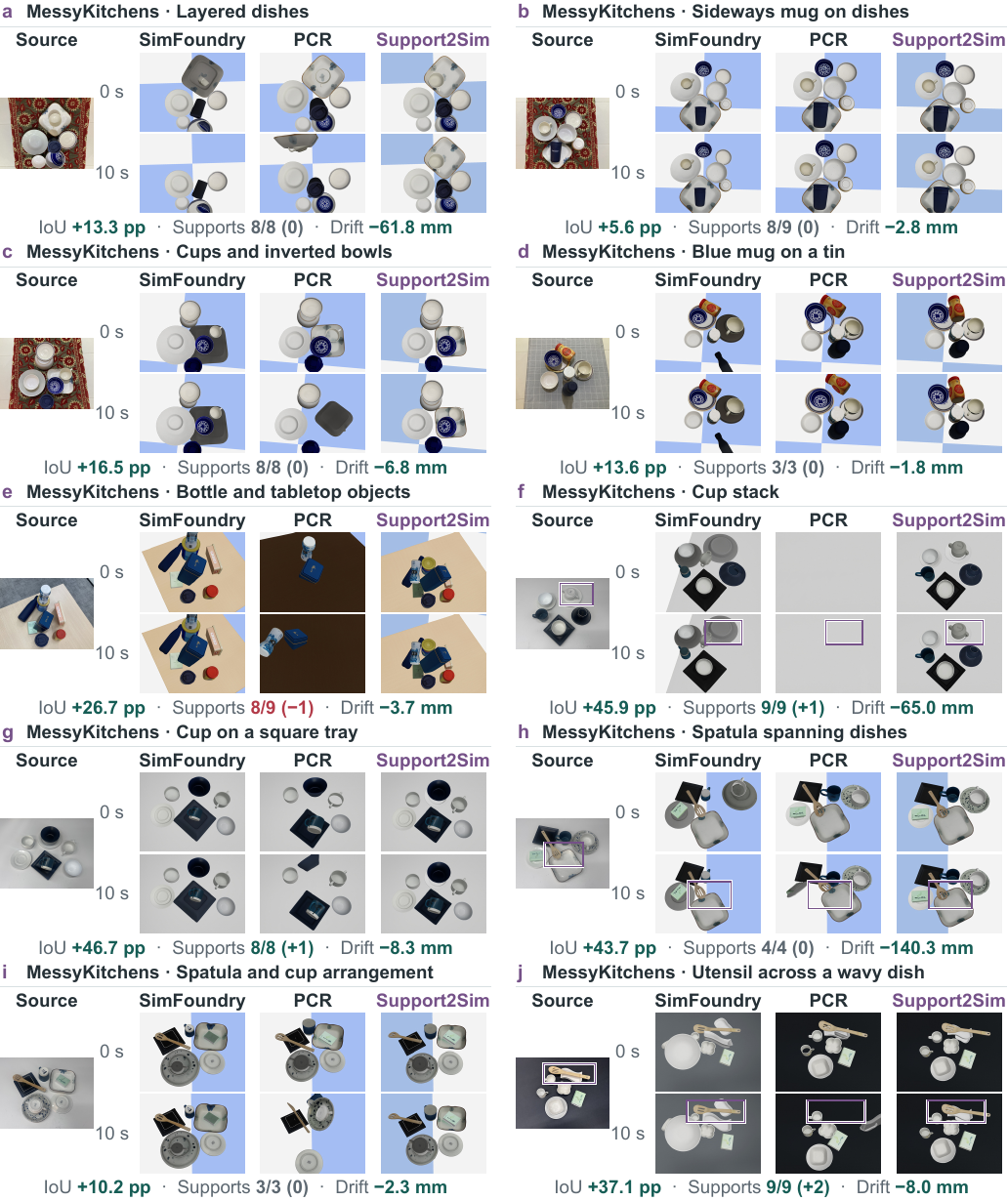}
  \caption{\textbf{Additional tabletop comparisons.}
  Source images and renders at release (upper) and 10 seconds (lower).
  Differences are Support2Sim minus the better of SimFoundry and PCR,
  chosen separately for each scene and metric: final IoU (percentage
  points), realized supports, and drift (mm). For example, $8/9\,(-1)$
  means eight of nine supports are realized, one fewer than the better
  baseline. Negative drift differences indicate improvement.
  Teal denotes gains, red regressions, and gray ties.}
  \label{fig:supp-qualitative-extended}
\end{figure*}

\begin{figure*}[p]
  \ContinuedFloat
  \centering
  \includegraphics[page=2,width=\textwidth]{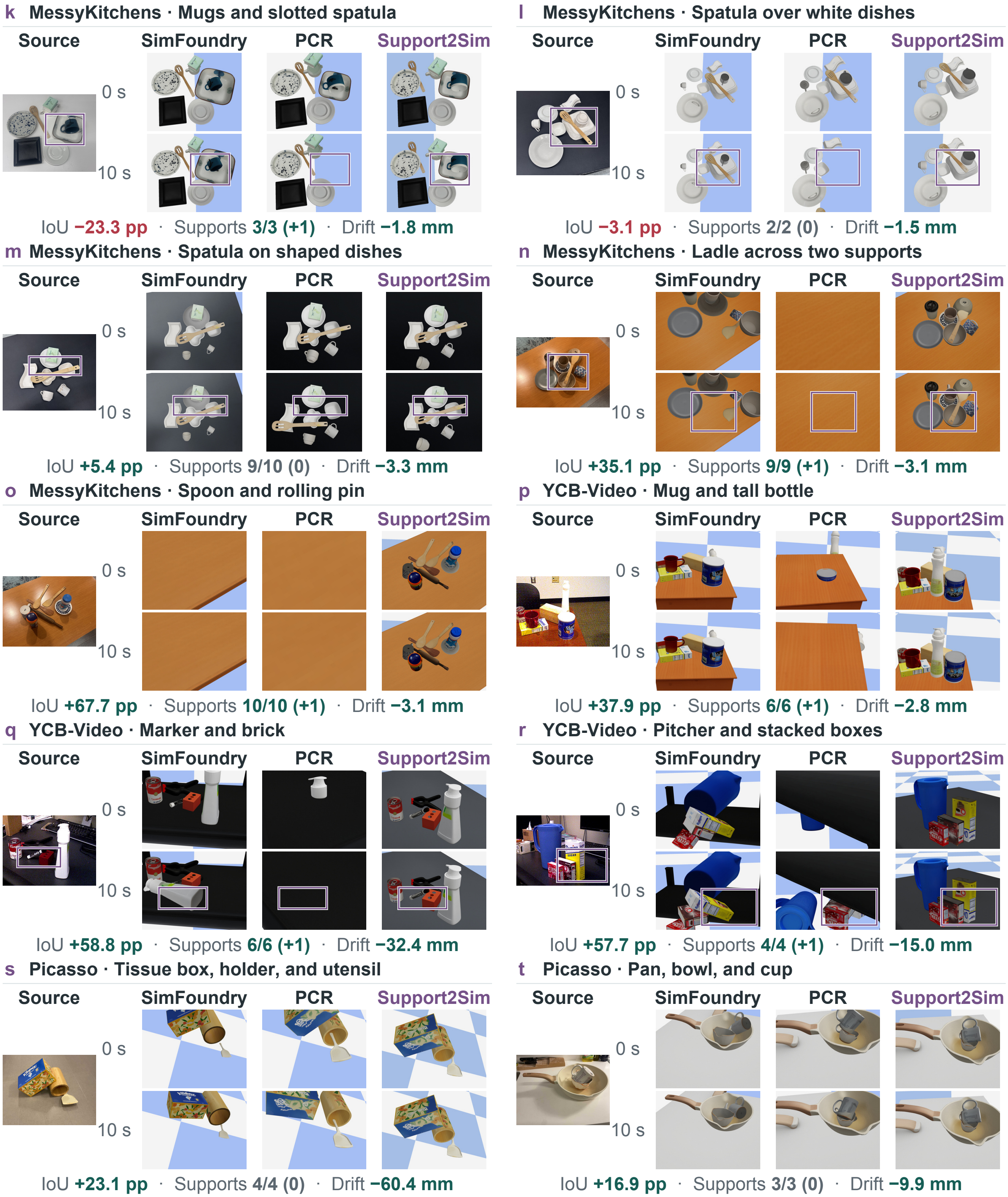}
  \caption{\textbf{Additional tabletop comparisons (continued).}
  The two pages show 20 additional scenes from MessyKitchens, YCB-Video,
  and Picasso using the same views and metric annotations.}
\end{figure*}

\begin{figure*}[p]
  \centering
  \includegraphics[width=\textwidth]{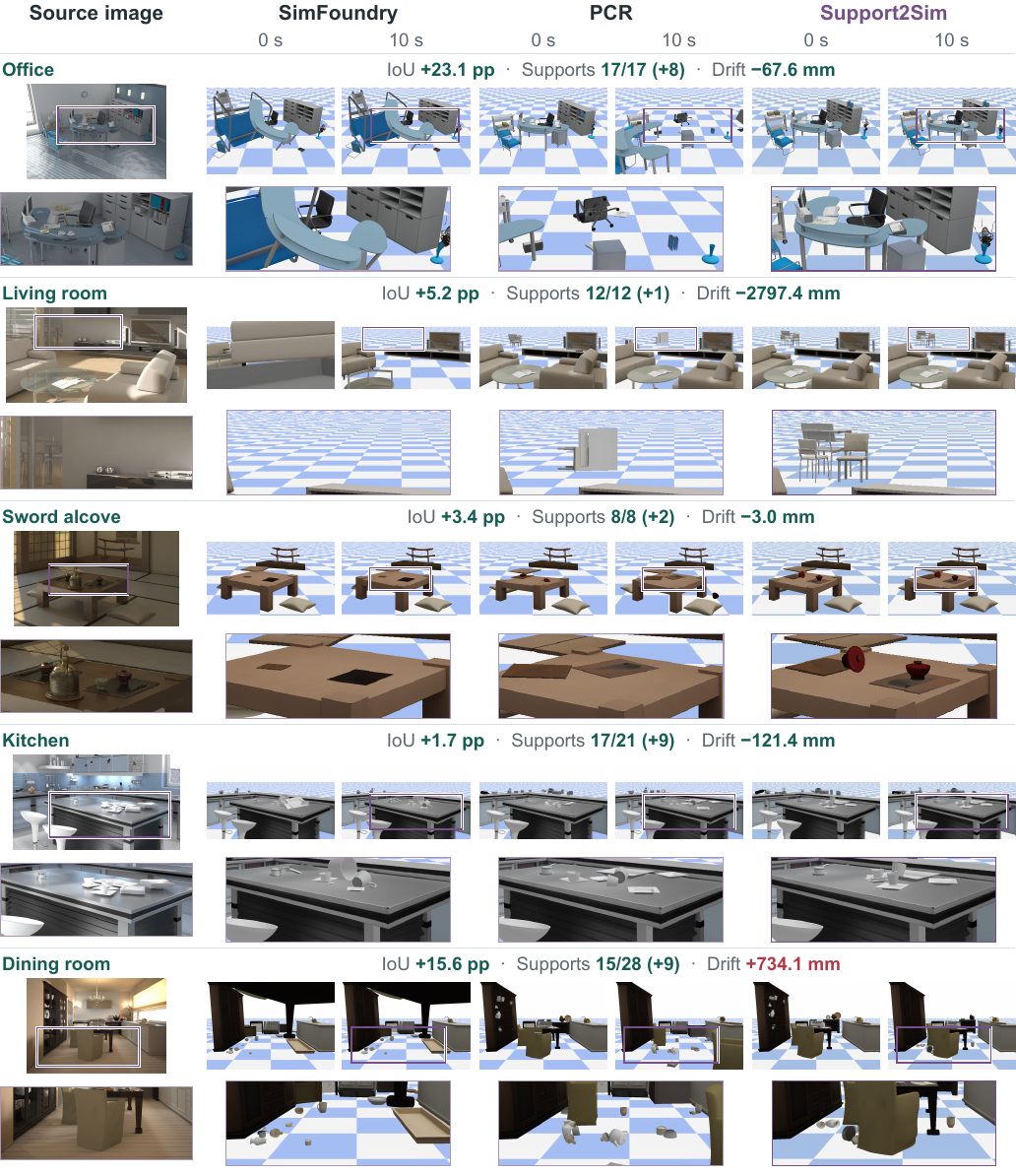}
  \caption{\textbf{Qualitative comparisons on all five HyperSim scenes.}
  Matched-camera renders at release and 10 seconds, with enlarged source
  and final-render crops below. Metric annotations follow
  Fig.~\ref{fig:supp-qualitative-extended}; the dining room retains visible
  failures.}
  \label{fig:supp-hypersim-fullpage}
\end{figure*}

\begin{figure*}[p]
  \centering
  \includegraphics[width=\textwidth]{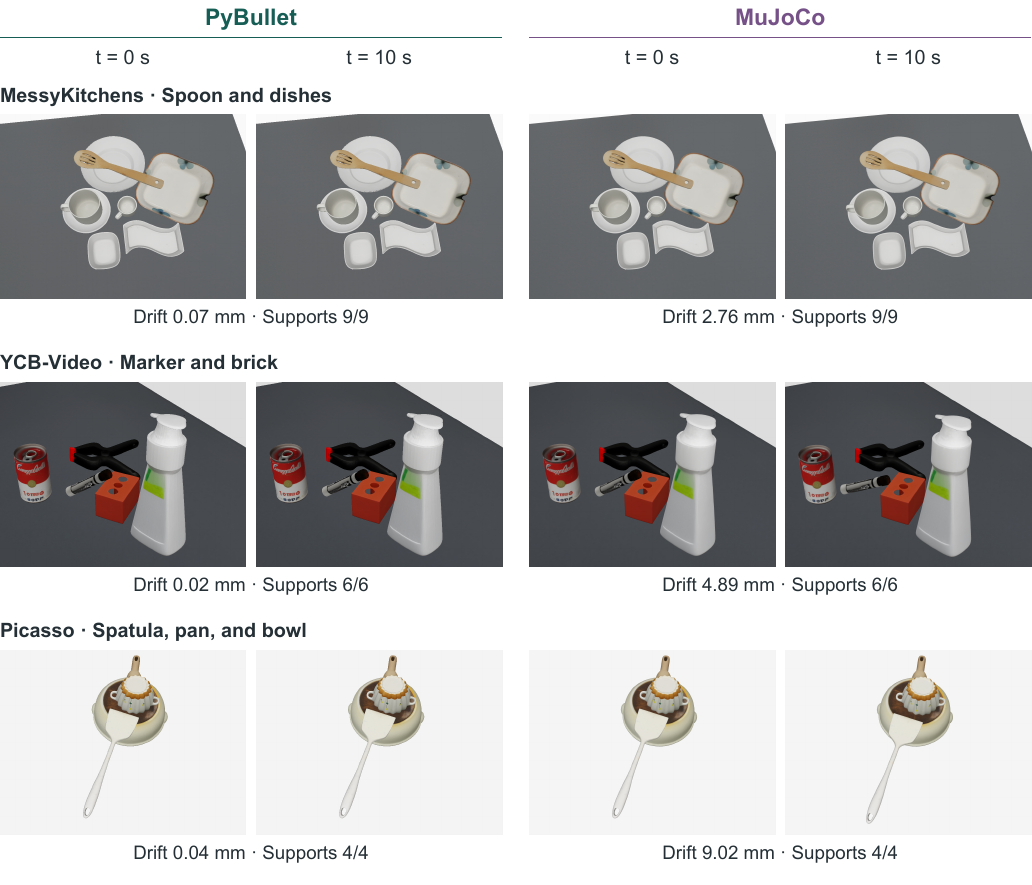}
  \caption{\textbf{Cross-simulator stability examples.}
  Three reconstructions at release and after 10 seconds in PyBullet and
  MuJoCo, starting from matching poses and physical properties.
  Cameras remain fixed. Annotations give mean object drift and realized
  intended supports; all supports are retained in these examples.}
  \label{fig:supp-simulator-exports}
\end{figure*}

\clearpage


\clearpage
\section{Policy Learning in Reconstructed Scenes}
\label{sec:supp-policy-learning}

We evaluate behavior cloning (BC), BC with DART noise injection
(BC + DART), and reinforcement learning (RL) with dense or sparse rewards
in four reconstructed YCB-Video scenes (Sec.~\ref{sec:policy-learning}).
We measure full-task success and intermediate milestones such as reaching
and lifting the object.

\subsection{Tasks and simulation setup}
\label{sec:supp-policy-setup}
Each task uses a Franka Panda to move one object while preserving the
surrounding arrangement (Table~\ref{tab:supp-policy-tasks}). We use
MuJoCo Playground environments with CPU MuJoCo~\cite{todorov2012mujoco,zakka2025mujocoplayground}.
We prepare copies of the reconstructions for the robot. We position each
scene within reach, correct task-specific object poses and centers of
mass, and configure the gripper. We replace the desk that is out of reach
with a lower support that has the same top outline.

Simulation uses MuJoCo 3.6.0 with a Newton solver, 50 iterations, and a
50\,Hz controller. The physics timestep is 2\,ms, except for banana
placement, which uses a 5\,ms timestep.

\noindent\textbf{Starting states and success.}
Arm joints are perturbed uniformly by $\pm0.05$\,rad, and the target object
is perturbed as listed in Table~\ref{tab:supp-policy-tasks}. Other objects
start at their settled poses. We allow 2 seconds of settling. If objects
overlap, the planned grasp is blocked, or neighboring objects have moved,
we reject the configuration and sample a new one.
RL uses the same randomization, with the gripper fully open at the start
of box transfer.
Success requires the target to reach its goal and be released, with no
finger-pad contact during the final second. During that second, its
horizontal position range must be at most 5\,mm, vertical range at most
3\,mm, tilt range at most $3^\circ$, and 0.2-second mean speed below
5\,cm/s. Other objects may move at most 1\,cm and tilt at most $15^\circ$;
for banana placement, the destination box must remain on its board with
tilt below $45^\circ$.

\subsection{Behavior cloning with and without DART}
\label{sec:supp-policy-bc}
A scripted expert approaches, grasps, lifts, transports, lowers, and
releases the target using inverse kinematics. It updates its plan from
the simulator state and retries failed grasps. Each BC model uses 3,000
successful demonstrations per task. Plain BC uses clean demonstrations;
BC + DART~\cite{laskey2017dart} adds Gaussian noise with standard deviation
0.3 to the seven arm actions in every second demonstration, leaving the
gripper unperturbed. The recorded label is the expert's clean action,
so the policy learns to recover from states reached under noisy control.

Policies observe simulator state, including joint positions and velocities,
object and gripper poses, and relative goal positions: 79 values per task,
or 92 for banana placement. They output seven joint-target increments and
one gripper increment at 50\,Hz. Arm increments are bounded by
0.04\,rad per control step. Training settings appear in
Table~\ref{tab:supp-policy-training}. Checkpoints are evaluated every five
epochs on 64 starts; selection maximizes success, breaking ties by lower
validation loss. Final BC + DART evaluation uses 128 fresh starts per task,
with deterministic policies and the full episode horizon.

\begin{table*}[t]
  \centering
  \small
  \setlength{\tabcolsep}{4pt}
  \renewcommand{\arraystretch}{1.12}
  \caption{\textbf{Robot-learning tasks.} Position and yaw perturbations are
  uniform. Horizons give the simulated duration of each evaluation episode.}
  \label{tab:supp-policy-tasks}
  \begin{tabular*}{\textwidth}{@{\extracolsep{\fill}}p{.14\textwidth}p{.43\textwidth}p{.22\textwidth}rr@{}}
    \toprule
    Task & Goal & Target perturbation & BC (s) & RL (s) \\
    \midrule
    Banana placement & Place the banana on the cracker box: bottom within
    5\,mm of its top and center of mass at least 3\,mm inside its footprint.
      & $\pm3$\,cm in $x,y$; $\pm20^\circ$ yaw & 25 & 28.1 \\
    Box stacking & Lay the gelatin box flat on the standing sugar box:
    bottom within 2\,mm of the top, center of mass at least 3\,mm inside,
    and thin axis within $15^\circ$ of vertical.
      & $\pm3$\,cm in $x,y$; $\pm20^\circ$ yaw & 30 & 21.1 \\
    Marker relocation & Move the leaning marker to a 4\,cm-radius goal
    disc beyond the soup can; its long axis must be within $15^\circ$ of
    horizontal, touching only the table.
      & $\pm1$\,cm along the block face; $\pm5^\circ$ yaw & 38 & 25.1 \\
    Box transfer & Move the sugar box from the inverted bowl to a
    4\,cm-radius goal disc on the support, lying flat at the target
    orientation ($20^\circ$). Up to 4\,cm of edge overhang is permitted.
      & $\pm1$\,cm in $x,y$; $\pm10^\circ$ yaw & 24 & 17.1 \\
    \bottomrule
  \end{tabular*}
\end{table*}

\begin{table*}[t]
  \centering
  \small
  \setlength{\tabcolsep}{4pt}
  \renewcommand{\arraystretch}{1.08}
  \caption{\textbf{Policy-training settings.} Dense and sparse RL share
  these settings. BC and PPO use different action interfaces and training
  budgets; the reward comparison is controlled within PPO.}
  \label{tab:supp-policy-training}
  \begin{tabular*}{\textwidth}{@{\extracolsep{\fill}}p{.24\textwidth}p{.35\textwidth}p{.35\textwidth}@{}}
    \toprule
    Setting & BC / BC + DART & Dense / sparse RL \\
    \midrule
    Network & 3 hidden layers of 256; SiLU; tanh output
      & Separate actor and critic; 2 hidden layers of 256; tanh \\
    Observation normalization & Training-set mean and standard deviation
      & Running observation and reward statistics \\
    Actions and frequency & 7 joint increments + gripper; 50\,Hz
      & 3 hand translations + 3 rotations + gripper; 10\,Hz \\
    Training budget & 3,000 successful demonstrations; at most 120 epochs
      & 3,004,416 policy steps from random initialization \\
    Batch / rollout & Batch size 4,096; 10\% of episodes held out
      & 6 environments $\times$ 1,024 steps per update \\
    PPO updates & --- & Minibatches of 2,048; 10 epochs per update \\
    Learning rate & $10^{-3}$ after 1,000 warm-up steps; cosine decay to $10^{-5}$
      & $3\times10^{-4}$ \\
    Objective settings & Mean squared action error; gripper weight 5
      & $\gamma=0.99$; GAE $\lambda=0.95$; clip 0.2 \\
    Optimization checks & Adam; gradient-norm limit 1; validation patience 8 epochs
      & Target KL 0.02; entropy coefficient 0 \\
    Hardware per run & One NVIDIA V100 GPU
      & 7 Intel Xeon E5-2698 v4 CPU cores \\
    \bottomrule
  \end{tabular*}
\end{table*}

\subsection{Dense and sparse reinforcement learning}
\label{sec:supp-policy-rl}
We train both RL variants from random initialization using proximal
policy optimization (PPO)~\cite{schulman2017ppo} in
Stable-Baselines3~\cite{raffin2021stablebaselines3}. For each task, only the
reward differs between the dense and sparse runs. The policy commands
translations of up to 2\,cm, rotations of up to $10^\circ$, and whether
to open or close the gripper. Inverse kinematics converts the hand
motion to joint targets.
Each action lasts five controller steps. After a grasp, the gripper stays
closed unless explicitly commanded to release. Arm exploration standard
deviation starts at 0.61 with a floor of 0.15; the gripper floor is 0.5.

\noindent\textbf{Rewards and training starts.}
Dense reward increases through approach, alignment, grasp, lift, transport
over the goal, release, and success. The respective stage scores are
0--1.2, approximately 1.8--3.85, 4--5, 5--5.9, 6--7, 7--8, and 10 before
penalties. Within a stage, distance and orientation terms reward progress;
closing is rewarded only near the grasp point with the object between the
fingers. Transport is rewarded after the object clears the goal surface.
Penalties discourage disturbing objects, abrupt actions, and release from
height. The score is divided by 10 before reward normalization.
The sparse reward is $r_t=\mathbf{1}[t=T]\mathbf{1}[\mathrm{success}(s_T)]$:
one only for success at the final step, and zero otherwise, with no shaping
or penalties.

Both RL variants use one grasp pose per object to define the grasp
geometry and set up training starts. We average this pose over 16 clean
expert rollouts. PPO receives no expert action labels or BC initialization.
Training starts are 60\% ordinary scene starts, 15\% with the hand above
the grasp point, 15\% already grasping the target, and 10\% holding it
above the goal. Evaluation always uses ordinary scene starts.

\noindent\textbf{Evaluation.}
For each task, we train each RL variant for 3,004,416 policy steps.
Every 100,000 steps, we evaluate on 16 fixed ordinary starts. We choose
the checkpoint with the highest success rate, using the dense evaluation
return to break ties. We then test it deterministically on 128 fresh
starts. Full-task success uses the same placement criteria as BC.
A separate \emph{lift-success} metric requires a sustained grasp with the
object's lowest point at least 5\,cm above its starting support for
1 second. This is stricter than the reward's 4\,cm lift stage.

\subsection{Performance and qualitative results}
\label{sec:supp-policy-results}
BC + DART succeeds on all four tasks, with a mean success rate of $91.0\%$.
Plain BC achieves at most $3.1\%$ success on any task
(Fig.~\ref{fig:supp-policy-performance}). Perturbed demonstrations give
the policy examples of how to recover from small deviations.
Figure~\ref{fig:supp-policy-rl-curves} tracks progress through individual
task stages. Dense RL learns to reach all four targets and to grasp and
lift the banana; in the other three tasks, it stops before grasping.
Sparse RL does not reach any target from ordinary starts. Dense rewards
provide feedback for these intermediate stages. In the final evaluation,
dense RL achieves sustained banana lifting on $119/128$ trials ($93.0\%$),
but does not complete transport and placement. Neither dense RL nor
sparse RL completes a full task within the three-million-step budget
(Table~\ref{tab:supp-policy-performance}).
Figure~\ref{fig:supp-policy-rl-rollouts} compares representative rollouts
from all four methods on the four tasks. BC + DART completes the shown
placements, while plain BC can disturb surrounding objects and the PPO
policies stop before completing the task.

\begin{figure*}[p]
  \centering
  \captionsetup{skip=5pt}
  \includegraphics[width=.97\linewidth,height=.232\textheight,keepaspectratio]{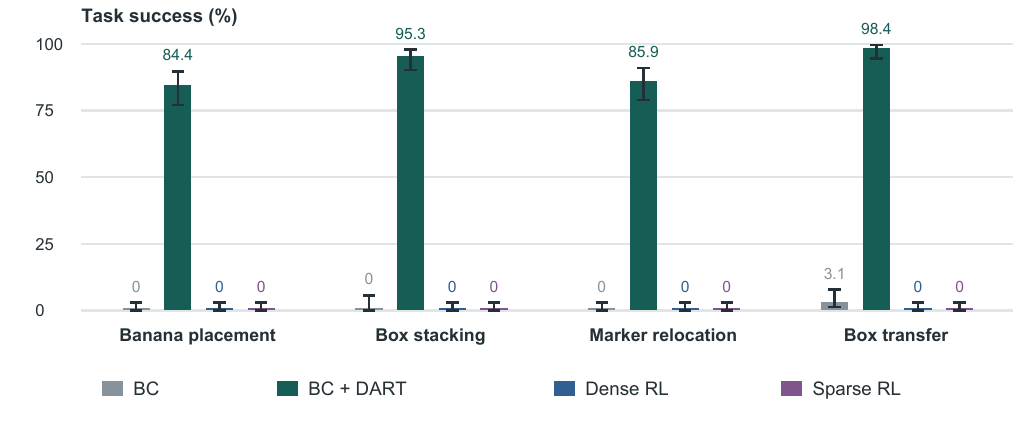}
  \caption{\textbf{Task completion across four learning methods.}
  Error bars are 95\% Wilson intervals over evaluation episodes.
  BC + DART learns successful placement in all four environments;
  dense and sparse RL have zero full-task successes within three million
  policy steps. Training protocols differ between BC and PPO; dense and
  sparse RL share the same setup apart from their reward.}
  \label{fig:supp-policy-performance}
  \par\smallskip
  \begin{minipage}{\linewidth}
    \centering
    \small
    \setlength{\tabcolsep}{5pt}
    \captionof{table}{\textbf{Policy-learning outcomes.} Entries count full-task
    successes out of all trials. Plain BC for box stacking used 64 trials
    with different seeds. All other evaluations used 128 trials.
    Bold teal marks the best full-task success for each task.}
    \label{tab:supp-policy-performance}
    \begin{tabular*}{\linewidth}{@{\extracolsep{\fill}}lcccc@{}}
      \toprule
      Task & BC & BC + DART & Dense RL & Sparse RL \\
      \midrule
      Banana placement & 0/128 & \textcolor[HTML]{165D55}{\textbf{108/128}} & 0/128 & 0/128 \\
      Box stacking & 0/64 & \textcolor[HTML]{165D55}{\textbf{122/128}} & 0/128 & 0/128 \\
      Marker relocation & 0/128 & \textcolor[HTML]{165D55}{\textbf{110/128}} & 0/128 & 0/128 \\
      Box transfer & 4/128 & \textcolor[HTML]{165D55}{\textbf{126/128}} & 0/128 & 0/128 \\
      \bottomrule
    \end{tabular*}
  \end{minipage}
  \par\smallskip
  \includegraphics[width=\linewidth,height=.46\textheight,keepaspectratio]{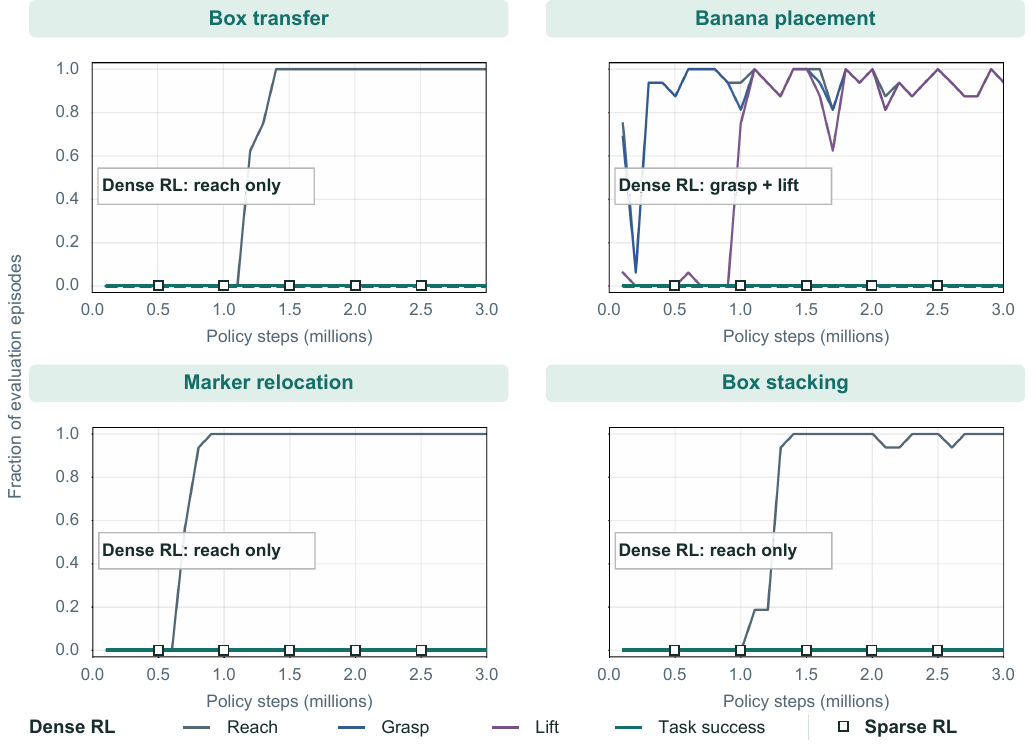}
  \caption{\textbf{Stages learned with dense and sparse RL.}
  Dense RL (solid colored curves) reaches the target in all four tasks
  and learns to grasp and lift the banana, but stops before grasping in
  the other three. Sparse RL (hollow black squares) reaches none of the
  stages. Full-task success remains zero for both within three million
  policy steps. Values are fractions of 16 fixed evaluation starts per
  task. Lift requires 4\,cm clearance; the final lift-success metric uses
  5\,cm for 1 second.}
  \label{fig:supp-policy-rl-curves}
\end{figure*}

\begin{figure*}[p]
  \centering
  \includegraphics[page=1,width=.85\textwidth]{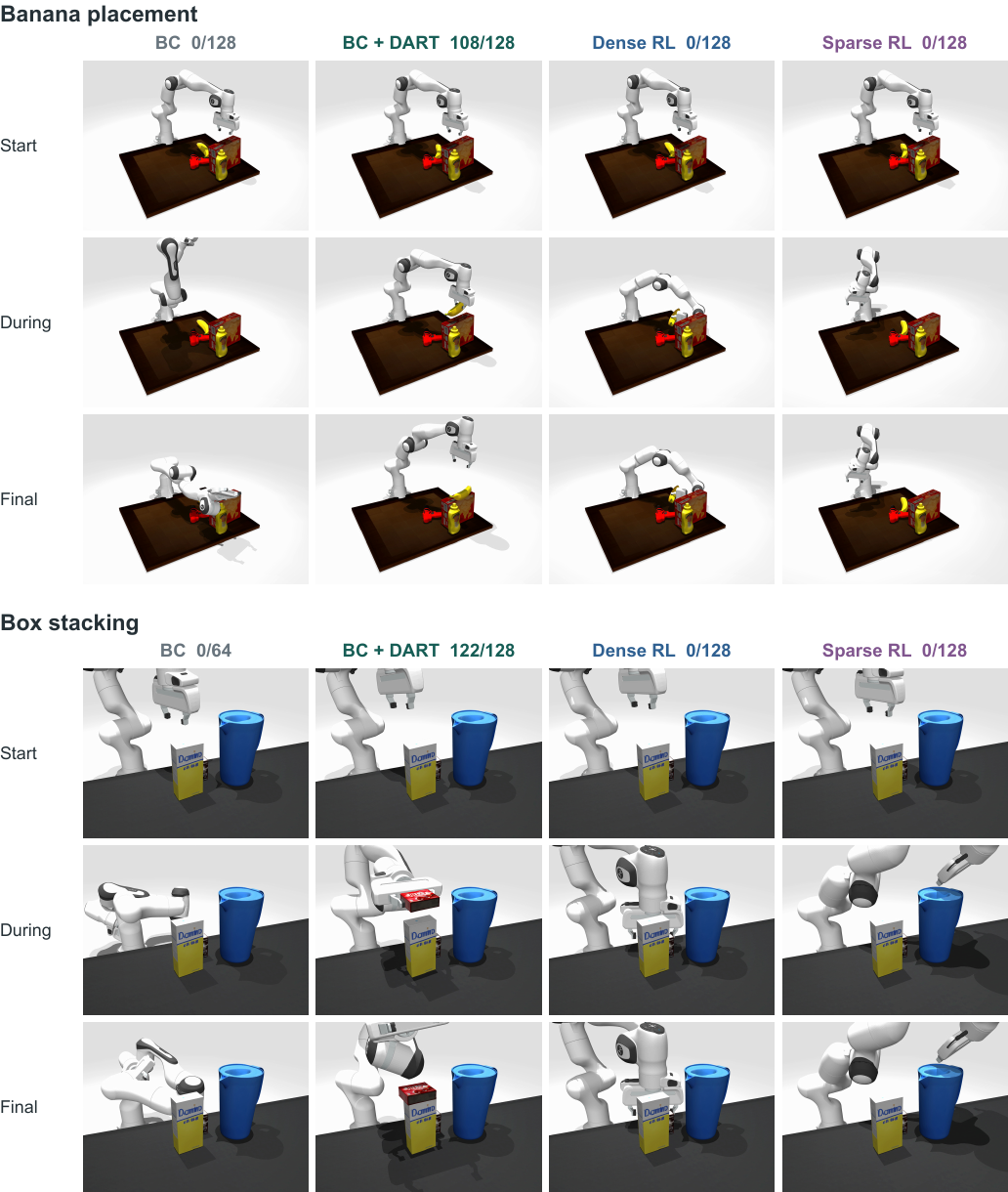}
  \caption{\textbf{Four-method policy-rollout comparison.}
  Banana placement and box stacking. Columns compare BC, BC + DART,
  dense RL, and sparse RL; rows show start, intermediate, and final
  states. Counts are full-task successes/trials. Frames come from separate
  episodes and are not time-synchronized.}
  \label{fig:supp-policy-rl-rollouts}
\end{figure*}

\begin{figure*}[p]
  \ContinuedFloat
  \centering
  \includegraphics[page=2,width=.85\textwidth]{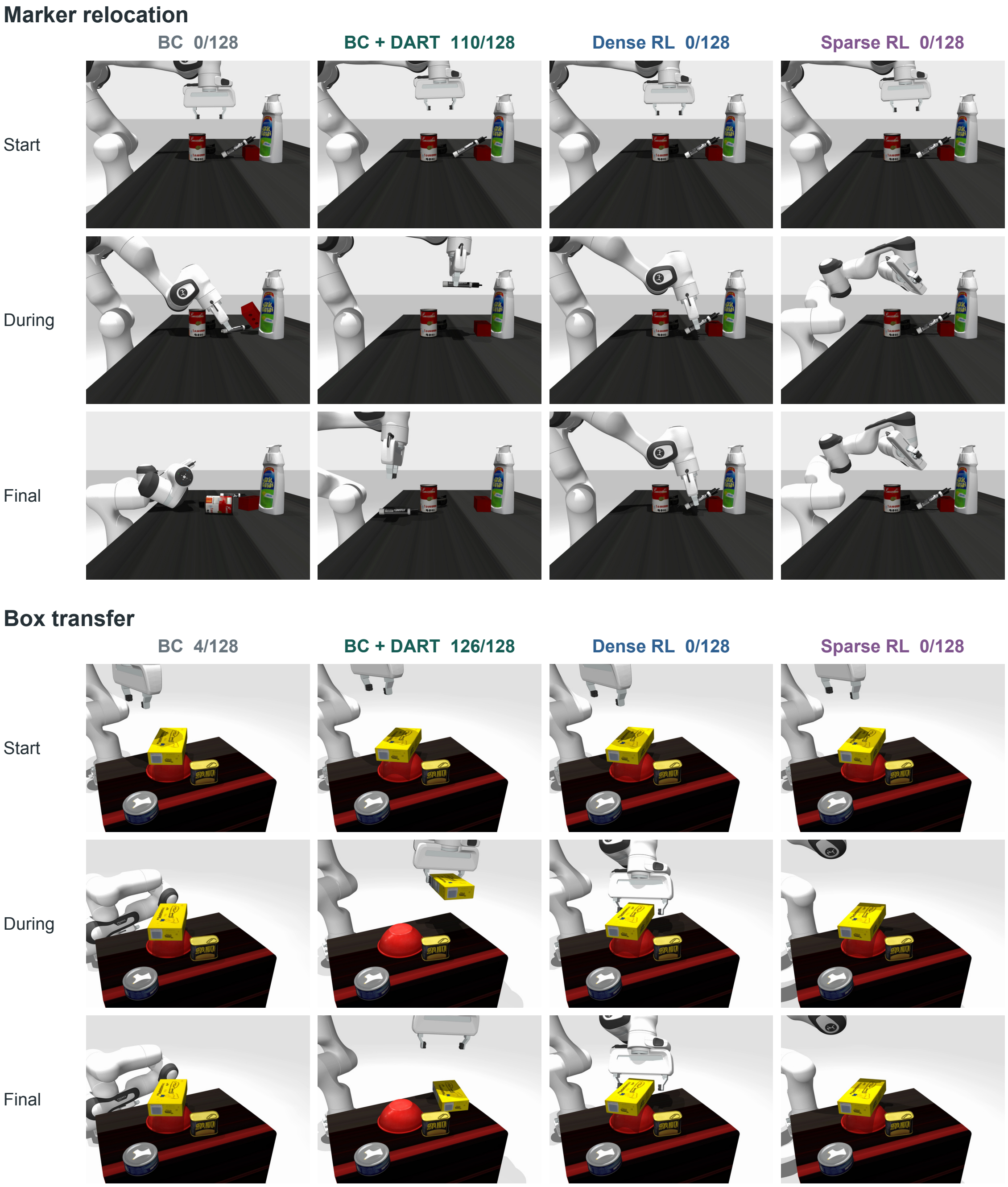}
  \caption{\textbf{Four-method policy-rollout comparison (continued).}
  Marker relocation and box transfer, with the same layout and counts.
  The marker's side view shows BC disturbing neighboring objects and
  BC + DART placing the marker beyond the can. All PPO examples use the
  final three-million-step runs.}
\end{figure*}

\clearpage

\newcommand{\Mesh}{\mathcal{M}}
%
\providecommand{\figdir}{figures/multiview/images}

\definecolor{accent}{HTML}{0E6B5E}      
\definecolor{accentsoft}{HTML}{E2F0EC}  
\definecolor{bluesoft}{HTML}{E4ECF7}
\definecolor{purplesoft}{HTML}{ECE6F5}
\definecolor{purplemid}{HTML}{6B4FA0}
\definecolor{warm}{HTML}{D9792B}        
\definecolor{warmsoft}{HTML}{FBE9D7}
\definecolor{ink}{HTML}{222222}
\definecolor{muted}{HTML}{6A6A6A}
\definecolor{rule}{HTML}{CFCFCF}
\definecolor{slotfill}{HTML}{F3F3F3}
\definecolor{slotline}{HTML}{A8A8A8}

\newcommand{\figfont}{\sffamily\fontsize{6.2}{7.2}\selectfont}
\newcommand{\figfonttiny}{\sffamily\fontsize{5}{5.8}\selectfont}
\newcommand{\slotfilefont}{\ttfamily\fontsize{4.2}{4.8}\selectfont}

\newsavebox{\slotbox}
\newcommand{\coverimage}[3]{%
  \sbox{\slotbox}{\includegraphics[width=100pt]{#3}}%
  \pgfmathsetmacro{\slotaspect}{100/(\ht\slotbox/1pt)}%
  \pgfmathsetmacro{\slottarget}{(#1)/(#2)}%
  \pgfmathparse{\slotaspect>\slottarget ? 1 : 0}%
  \ifnum\pgfmathresult=1\relax
    \includegraphics[height=#2]{#3}%
  \else
    \includegraphics[width=#1]{#3}%
  \fi
}

\tikzset{
  slotbase/.style={inner sep=0pt, outer sep=0pt, rounded corners=1.2pt},
  slotempty/.style={slotbase, draw=slotline, densely dashed, line width=0.35pt,
    fill=slotfill, align=center, text=muted, font=\figfonttiny},
  slotfull/.style={slotbase, draw=rule, line width=0.3pt},
  slotimage/.style={anchor=center, inner sep=0pt, outer sep=0pt, minimum width=0pt,
    minimum height=0pt, draw=none, fill=none, rounded corners=0pt, path picture={}},
}

\newcommand{\photoslot}[6][]{%
  \def\slotfile{}%
  \IfFileExists{\figdir/#5.jpg}{\def\slotfile{\figdir/#5.jpg}}{%
    \IfFileExists{\figdir/#5.png}{\def\slotfile{\figdir/#5.png}}{%
      \IfFileExists{\figdir/#5.pdf}{\def\slotfile{\figdir/#5.pdf}}{}}}%
  \ifx\slotfile\empty
    \node[slotempty, minimum width=#3, minimum height=#4, text width=#3-3pt, #1] (#2)
      {#6\\[-0.5pt]{\slotfilefont\detokenize{#5}}};
  \else
    \node[slotfull, minimum width=#3, minimum height=#4, #1,
      path picture={\node[slotimage] at (path picture bounding box.center)
        {\coverimage{#3}{#4}{\slotfile}};}] (#2) {};
  \fi
}

\newcommand{\photoslotmini}[6][]{%
  \def\slotfile{}%
  \IfFileExists{\figdir/#5.jpg}{\def\slotfile{\figdir/#5.jpg}}{%
    \IfFileExists{\figdir/#5.png}{\def\slotfile{\figdir/#5.png}}{%
      \IfFileExists{\figdir/#5.pdf}{\def\slotfile{\figdir/#5.pdf}}{}}}%
  \ifx\slotfile\empty
    \node[slotempty, minimum width=#3, minimum height=#4, text width=#3-2pt, #1] (#2) {#6};
  \else
    \node[slotfull, minimum width=#3, minimum height=#4, #1,
      path picture={\node[slotimage] at (path picture bounding box.center)
        {\coverimage{#3}{#4}{\slotfile}};}] (#2) {};
  \fi
}

\tikzset{
  badge/.style={font=\sffamily\fontsize{5.5}{6}\selectfont\bfseries, text=accent,
    fill=accentsoft, rounded corners=1pt, inner xsep=2.2pt, inner ysep=1.6pt},
  bandtitle/.style={font=\rmfamily\fontsize{9}{10}\selectfont, text=ink, anchor=west, inner sep=0pt},
  paneltitle/.style={font=\sffamily\fontsize{6.6}{7.4}\selectfont\bfseries, text=ink,
    anchor=base west, inner sep=0pt},
  lab/.style={font=\figfont, text=ink, inner sep=1pt},
  labmuted/.style={font=\figfonttiny, text=muted, inner sep=1pt},
  flow/.style={-{Triangle[length=3.2pt,width=3.2pt]}, draw=accent, line width=0.7pt},
  schem/.style={fill=accentsoft, rounded corners=1.5pt, draw=none},
  chip/.style={font=\figfonttiny, text=ink, fill=white, draw=rule, line width=0.3pt,
    rounded corners=1pt, inner xsep=2pt, inner ysep=1.4pt},
}

\renewcommand{\figdir}{figures/multiview/images}

\section{Applications: Multi-view Scene Reconstruction}
\label{sec:supp-multiview}

Similar to Sec.~\ref{sec:policy-learning}, we also present an application that attempts to extend the scale of the reconstructed scenes using multiple images. This application is presented as a tangential capability of our existing real2sim pipeline, where emphasis is placed on multi-view object association and reconstruction, rather than realized contact fidelity and scene stability.  

Given $N$ unposed RGB images $\{I_v\}_{v=1}^{N}$ of a static scene, we reconstruct a
gravity-aligned scene in which every object is a separate mesh with a
pose on its support (Fig.~\ref{fig:supp-mv-architecture}). A
generative image-to-3D model supplies each object's \emph{shape} from one
well-chosen view, and multi-view geometry supplies its \emph{pose}, and this helps alleviate ambiguities that can arise from using only a single view, such as occluded or hidden surfaces.

\begin{figure*}[tp]
  \centering
  \resizebox{\textwidth}{!}{\input{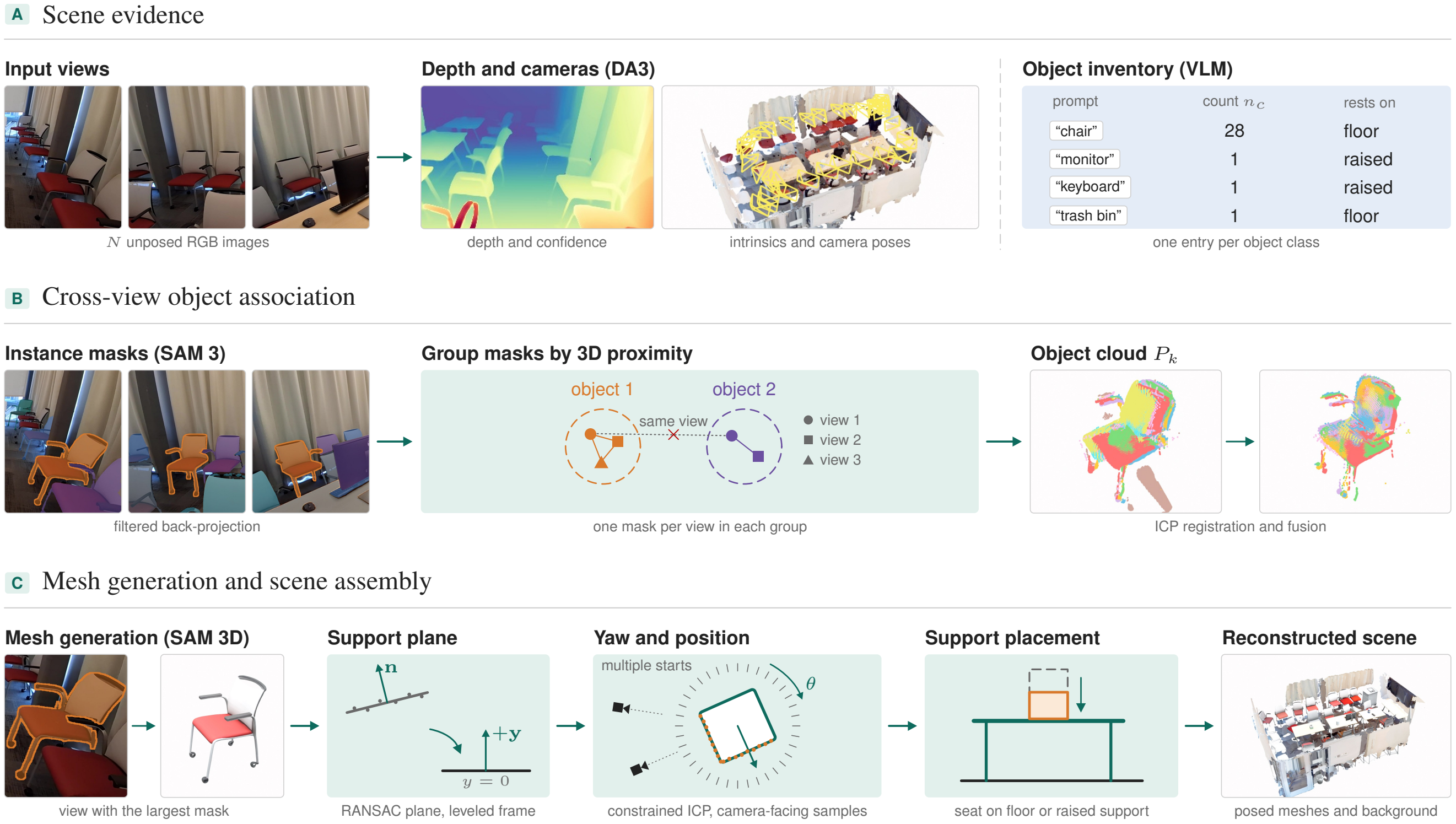}}
  \caption{\textbf{Multi-view scene reconstruction.}
  \textbf{(A)~Scene evidence.} DA3 estimates depth and camera parameters from
  unposed RGB images, and a vision--language model builds an object inventory
  with segmentation prompts, instance counts, and support types.
  \textbf{(B)~Cross-view object association.} We back-project SAM~3 masks,
  group them by 3D centroid proximity, keep at most one mask per view in each
  group, and register and fuse each object's clouds.
  \textbf{(C)~Mesh generation and scene assembly.} SAM~3D Objects generates an
  upright mesh from a selected view. We level the scene using the main support
  plane, fit each mesh's yaw and horizontal position to its object cloud,
  and seat the mesh on the floor or a raised support.}
  \label{fig:supp-mv-architecture}
\end{figure*}

\subsection{Method}
\label{sec:supp-mv-method}
We estimate scene geometry and compile an object inventory, then use them to
associate instance masks across views and fuse one point cloud per object
(Sec.~\ref{sec:supp-mv-evidence}). We generate each object's mesh from a
selected view and fit it to the fused observations before placing it on its
inferred support (Sec.~\ref{sec:supp-mv-assembly}).

\subsubsection{Scene evidence and cross-view data association}
\label{sec:supp-mv-evidence}
Depth Anything~3 (DA3)~\cite{lin2026depthanything3} jointly estimates metric
depth, confidence, intrinsics, and camera poses from the input views. The
vision--language model from Sec.~\ref{sec:scene-evidence} inventories object
classes and provides segmentation prompts, expected instance counts, and
support types (Fig.~\ref{fig:supp-mv-architecture}A).

SAM~3~\cite{carion2026sam3} segments each view using the inventory's class prompts
(Fig.~\ref{fig:supp-mv-architecture}B). We filter background depth from each
mask and back-project it into 3D. We group same-class masks by 3D centroid
proximity and allow at most one mask per view
in each group, so nearby objects seen together remain distinct. For example,
suppose view~1 shows two chairs A and B, and view~2 shows only A, and there is a 6\,cm distance between the two 3D centroids
of A from the two views and 28\,cm between the centroids of A in view~2 to B in view~1. We merge the two masks of A
first and reject the longer link between A in view~2 and B in view~1, because it would put both view~1 masks of chairs A and B in
one group. We then register each object's per-view clouds with
ICP~\cite{chen1992object} before fusing them into an object cloud $P_k$, reducing
the spread from residual depth errors.

\subsubsection{Mesh generation and scene assembly}
\label{sec:supp-mv-assembly}
For each object, we select the view with the largest mask, preferring masks
fully inside the image. SAM~3D Objects~\cite{chen2026sam3d} uses that view's
image, mask, and DA3 pointmap to generate an upright canonical mesh $\Mesh_k$
(Fig.~\ref{fig:supp-mv-architecture}C).

To place the meshes in a common frame, we fit a dominant plane to the fused
scene cloud with RANSAC~\cite{fischler1981random}. We accept it when its
normal is close to the reference camera's up direction, which avoids treating
a wall as the main support. Otherwise, we use a horizontal plane at the
median base height of the object clouds. We level the scene using the support
normal and set the plane height to zero.

Starting from several evenly spaced yaw angles, we run constrained ICP to match
observed points to mesh samples, updating only yaw and horizontal position.
We rank the resulting poses by how well the mesh fits the fused object cloud,
with an additional penalty when mesh samples whose outward normals facing an
observing camera lie far from the cloud.

We seat floor-supported meshes on this plane when the lower points of their
clouds lie near it, allowing for partially occluded bases. If the object
inventory indicates a raised support and the observed base lies above the
plane, we look for a reconstructed support mesh below the object, such as a
table.
We use its local top height when it is close to the object's observed base.
If no suitable mesh is available, we use the highest broad, flat layer of
scene points under the object, falling back to the cloud's base height if
no such layer is found. Seating changes only the vertical position.

We remove each object's original points from the fused scene cloud and
insert its posed mesh. The result contains separate object meshes and the
remaining background geometry in a common gravity-aligned frame
(Fig.~\ref{fig:supp-mv-results}).

\subsection{Results}
\label{sec:supp-mv-results}
We present qualitative results for multi-view scene extension in four scenes
(Fig.~\ref{fig:supp-mv-results}). We recorded a video of each scene and sampled
frames at intervals of 1.5\,seconds on average across the four scenes to obtain the input
views. To describe the total mapped extent, we report the \emph{scene extent}:
the length and width of the smallest ground-plane rectangle that encloses the
fused DA3 scene cloud, ignoring isolated points.

\paragraph{Conference room.}
The scene extent is 7.0\,m $\times$ 5.0\,m. A 54\,second video provides 36 views
along a 16.5\,m loop around the room. A row of four tables runs down the
middle, with chairs around them and along the walls. The reconstruction
contains 43 separate object meshes: 27 chairs, the four tables, and 12 other
objects ranging from a whiteboard and a cabinet to a keyboard and a remote
control. Eight of these rest on a table or on the cabinet, not on the floor.

\paragraph{Potted plants in a lab.}
The scene extent is 16.3\,m $\times$ 9.3\,m. This is the shortest capture: a
15\,second video provides 10 views along a straight 9.4\,m walk. Ten potted plants,
each about 0.8\,m tall, stand in two rows of five, with the rows 1.8\,m apart
and the plants about 2.2\,m apart along each row. They occupy a
9.5\,m $\times$ 2.6\,m strip of the mapped lab space. Each plant is segmented in
only two to five views, and all ten are reconstructed as separate meshes.

\paragraph{Outdoor seating area.}
The scene extent is 96.0\,m $\times$ 61.7\,m, the largest of the four. A
179\,second video provides 96 views along a 116\,m walk. The extent includes
distant, low-confidence points on the surroundings; with these removed, as in
Fig.~\ref{fig:supp-mv-results}, the cloud covers about 44\,m $\times$ 20\,m.
The seating area holds two rows of table and chair sets, about 9\,m apart and 30\,m
long. The reconstruction contains 57 objects: 39 chairs and 18 tables. The
furniture is thin and partly see-through, and most objects are segmented in
only one or two views.

\paragraph{Atrium and corridors.}
The scene extent is 45.8\,m $\times$ 38.7\,m. This is the longest capture: a
233\,second video provides 154 views along a 139\,m walk that crosses the atrium, loops through the corridors
behind it, and returns to the atrium. The extent includes distant,
low-confidence points beyond the walked area; with these removed, as in
Fig.~\ref{fig:supp-mv-results}, the cloud covers about 40\,m $\times$ 34\,m.

A single DA3 solve over all 154 views recovers the atrium but loses track in
the narrow, plain-walled corridors, where consecutive views share little
similarities: 30 of the camera steps there jump by 2.5\,m to 8\,m. We therefore
resample the corridor stretch of the video more densely, about every 0.5\,s,
and solve it as six overlapping chunks of 63 to 85 frames, each of which is
consistent on its own. We chain the chunks with similarity transforms
estimated from the 3D points of the frames that neighboring chunks share,
and fix both ends of the chain to the atrium views of the full solve. 

We then level each chunk's floor onto the atrium floor. The corridor views
take their depth and poses from the stitched chunks, and the remaining views
keep those of the full solve.

The reconstruction contains 125 objects in 18 classes, including 25 chairs,
16 sofas and seating booths with 9 pillows, 10 tables, 13 trash bins, and
7 potted plants. They form a work area of tables with chairs and lounges of
sofas around low tables in the atrium, and seating booths, bins, and plants
along the corridors.

\begin{figure*}[p]
  \centering
  \def\mvresultspage{1}%
  \resizebox{\textwidth}{!}{
\begin{tikzpicture}[x=1cm, y=1cm, line cap=round, line join=round]
  \providecommand{\mvresultspage}{1}
  \def\figW{17.4}
  \def\rowh{3.6}         
  \def\closeh{4.13}      
  \def\titleh{0.40}      
  \def\blockpitch{8.93}  
  \def\xin{0.62}         
  \def\gap{0.12}
  \def\win{3.66}         
  \def\wsc{4.83}         
  \pgfmathsetmacro{\xsc}{\xin+\win+\gap}
  \pgfmathsetmacro{\xsb}{\xsc+\wsc+\gap}         
  \pgfmathsetmacro{\xtp}{\xsb+\wsc+\gap}
  \pgfmathsetmacro{\wtp}{\figW-\xtp}             
  \pgfmathsetmacro{\thumbw}{(\win-0.06)/2}
  \pgfmathsetmacro{\thumbh}{(\rowh-0.06)/2}
  \pgfmathsetmacro{\wcl}{(\figW-\xin-2*\gap)/3}
  \pgfmathsetmacro{\ycl}{-\rowh-\titleh}
  \pgfmathsetmacro{\blockh}{\rowh+\titleh+\closeh}

  \newcommand{\sceneblock}[4]{%
    \begin{scope}[yshift={(-#1*\blockpitch)*1cm}]
      \foreach \x/\w/\t in {\xin/\win/Input views,
                            \xsc/\wsc/Reconstruction{,} view 1, \xsb/\wsc/Reconstruction{,} view 2,
                            \xtp/\wtp/Top-down layout} {
        \node[paneltitle, anchor=base] at ({\x+\w/2},0.24) {\t};
      }
      \draw[rule, line width=0.4pt] (0,0.14) -- (\figW,0.14);
      \node[rotate=90, anchor=center, font=\sffamily\fontsize{7}{8}\selectfont\bfseries, text=ink]
        at (0.14,{-\blockh/2}) {#3};
      \node[rotate=90, anchor=center, labmuted] at (0.42,{-\blockh/2}) {#4};
      \foreach \i/\cx/\cy in {1/0/0, 2/1/0, 3/0/1, 4/1/1} {
        \photoslotmini[anchor=north west,
          at={({\xin+\cx*(\thumbw+0.06)},{-\cy*(\thumbh+0.06)})}]
          {#2v\i}{\thumbw cm}{\thumbh cm}{res_#2_view\i}{view \i}
      }
      \photoslot[anchor=north west, at={(\xsc,0)}]{#2sc}{\wsc cm}{\rowh cm}{res_#2_scene}{reconstruction, view 1}
      \photoslot[anchor=north west, at={(\xsb,0)}]{#2sb}{\wsc cm}{\rowh cm}{res_#2_scene2}{reconstruction, view 2}
      \photoslot[anchor=north west, at={(\xtp,0)}]{#2tp}{\wtp cm}{\rowh cm}{res_#2_top}{reconstruction, top-down}
      \node[paneltitle] at (\xin,{\ycl+0.12}) {Close-up views};
      \foreach \i in {1,2,3} {
        \photoslot[anchor=north west, at={({\xin+(\i-1)*(\wcl+\gap)},\ycl)}]
          {#2c\i}{\wcl cm}{\closeh cm}{res_#2_close\i}{reconstruction, close-up \i}
      }
    \end{scope}
  }

  \ifnum\mvresultspage=1\relax
    \sceneblock{0}{room}{Meeting room}{36 views, 43 objects}
    \sceneblock{1}{plants}{Potted plants}{10 views, 10 objects}
  \else
    \sceneblock{0}{chairs}{Outdoor seating}{96 views, 57 objects}
    \sceneblock{1}{atrium}{Atrium and corridors}{154 views, 125 objects}
  \fi
\end{tikzpicture}}
  \caption{\textbf{Reconstructed scenes.} For each scene, the upper row shows
  four of the input views and our reconstruction from two perspective views
  and a top-down view, and the lower row shows three close-up views of the
  same reconstruction. This page shows an indoor meeting room and a group of
  potted plants. For display, ceilings and other high points are cut away,
  low-confidence DA3 points are hidden, and in some close-up views the
  background between the camera and the objects is cut away.}
  \label{fig:supp-mv-results}
\end{figure*}
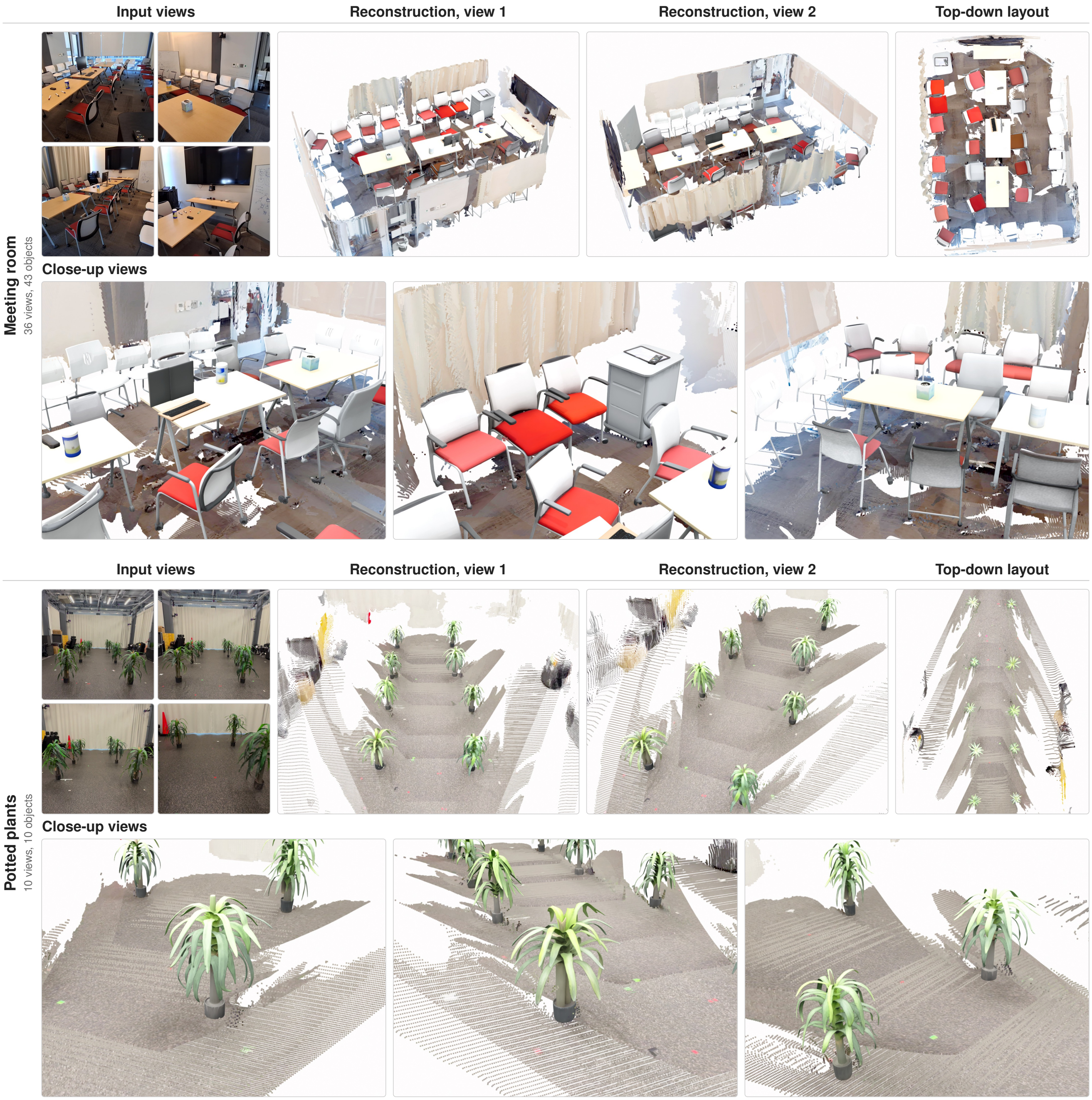

\begin{figure*}[p]
  \ContinuedFloat
  \centering
  \def\mvresultspage{2}%
  \resizebox{\textwidth}{!}{
\begin{tikzpicture}[x=1cm, y=1cm, line cap=round, line join=round]
  \providecommand{\mvresultspage}{1}
  \def\figW{17.4}
  \def\rowh{3.6}         
  \def\closeh{4.13}      
  \def\titleh{0.40}      
  \def\blockpitch{8.93}  
  \def\xin{0.62}         
  \def\gap{0.12}
  \def\win{3.66}         
  \def\wsc{4.83}         
  \pgfmathsetmacro{\xsc}{\xin+\win+\gap}
  \pgfmathsetmacro{\xsb}{\xsc+\wsc+\gap}         
  \pgfmathsetmacro{\xtp}{\xsb+\wsc+\gap}
  \pgfmathsetmacro{\wtp}{\figW-\xtp}             
  \pgfmathsetmacro{\thumbw}{(\win-0.06)/2}
  \pgfmathsetmacro{\thumbh}{(\rowh-0.06)/2}
  \pgfmathsetmacro{\wcl}{(\figW-\xin-2*\gap)/3}
  \pgfmathsetmacro{\ycl}{-\rowh-\titleh}
  \pgfmathsetmacro{\blockh}{\rowh+\titleh+\closeh}

  \newcommand{\sceneblock}[4]{%
    \begin{scope}[yshift={(-#1*\blockpitch)*1cm}]
      \foreach \x/\w/\t in {\xin/\win/Input views,
                            \xsc/\wsc/Reconstruction{,} view 1, \xsb/\wsc/Reconstruction{,} view 2,
                            \xtp/\wtp/Top-down layout} {
        \node[paneltitle, anchor=base] at ({\x+\w/2},0.24) {\t};
      }
      \draw[rule, line width=0.4pt] (0,0.14) -- (\figW,0.14);
      \node[rotate=90, anchor=center, font=\sffamily\fontsize{7}{8}\selectfont\bfseries, text=ink]
        at (0.14,{-\blockh/2}) {#3};
      \node[rotate=90, anchor=center, labmuted] at (0.42,{-\blockh/2}) {#4};
      \foreach \i/\cx/\cy in {1/0/0, 2/1/0, 3/0/1, 4/1/1} {
        \photoslotmini[anchor=north west,
          at={({\xin+\cx*(\thumbw+0.06)},{-\cy*(\thumbh+0.06)})}]
          {#2v\i}{\thumbw cm}{\thumbh cm}{res_#2_view\i}{view \i}
      }
      \photoslot[anchor=north west, at={(\xsc,0)}]{#2sc}{\wsc cm}{\rowh cm}{res_#2_scene}{reconstruction, view 1}
      \photoslot[anchor=north west, at={(\xsb,0)}]{#2sb}{\wsc cm}{\rowh cm}{res_#2_scene2}{reconstruction, view 2}
      \photoslot[anchor=north west, at={(\xtp,0)}]{#2tp}{\wtp cm}{\rowh cm}{res_#2_top}{reconstruction, top-down}
      \node[paneltitle] at (\xin,{\ycl+0.12}) {Close-up views};
      \foreach \i in {1,2,3} {
        \photoslot[anchor=north west, at={({\xin+(\i-1)*(\wcl+\gap)},\ycl)}]
          {#2c\i}{\wcl cm}{\closeh cm}{res_#2_close\i}{reconstruction, close-up \i}
      }
    \end{scope}
  }

  \ifnum\mvresultspage=1\relax
    \sceneblock{0}{room}{Meeting room}{36 views, 43 objects}
    \sceneblock{1}{plants}{Potted plants}{10 views, 10 objects}
  \else
    \sceneblock{0}{chairs}{Outdoor seating}{96 views, 57 objects}
    \sceneblock{1}{atrium}{Atrium and corridors}{154 views, 125 objects}
  \fi
\end{tikzpicture}}
  \caption{\textbf{Reconstructed scenes (continued).} An outdoor seating area
  and an atrium with lounge and work areas and the corridors behind it, in the
  same layout. For these two
  larger scenes, the perspective views show parts of the scene.}
\end{figure*}
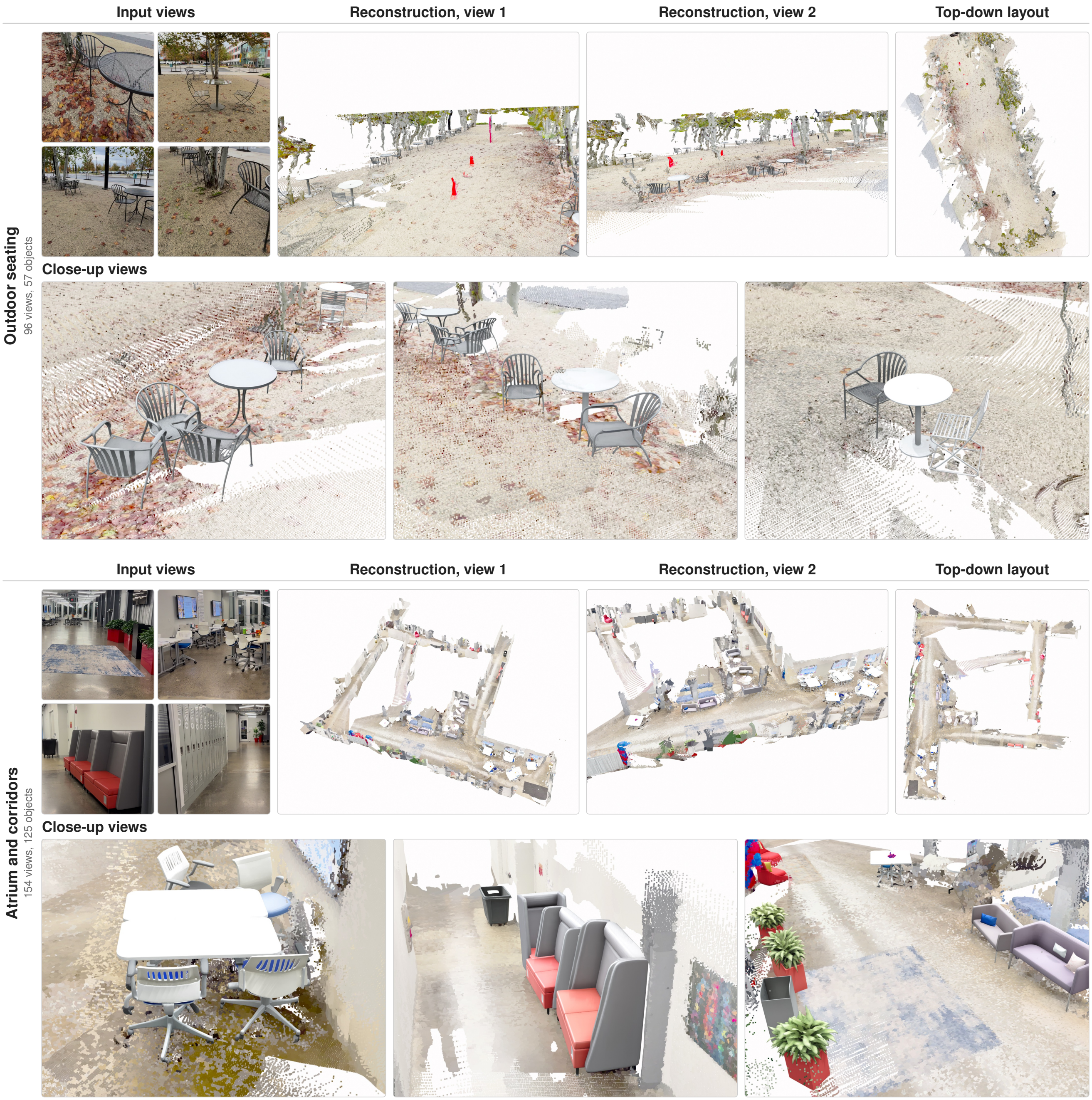

\clearpage
\endgroup
\end{document}